# GPT versus Humans – Uncovering Ethical Concerns in Conversational Generative AI-empowered Multi-Robot Systems

**Authors:**

1.*Rebekah Rousi[1],
Rebekah.rousi@uwasa.fi

2. Niko Mäkitalo[2],
niko.k.makitalo@jyu.fi

3. Hooman Samani[3],
h.samani@arts.ac.uk

4. Kai-Kristian Kemell[4],
kai-kristian.kemell@helsinki.fi

5. Jose Siqueira de Cerqueira[5],
jose.siqueiradecerqueira@tuni.fi

6. Ville Vakkuri[1],
ville.vakkuri@uwasa.fi

7. Tommi Mikkonen[2],
tommi.j.mikkonen@jyu.fi

8. Pekka Abrahamsson[5],
pekka.abrahamsson@tuni.fi

*Corresponding author
[1]School of Marketing & Communication University of Vaasa, Finland
[2] Faculty of Information Technology University of Jyväskylä, Finland
[3]Creative Robotics Creative Computing Institute University of the Arts London, UK
[4]Computer and Information Sciences University of Helsinki, Finland
[5]Faculty of Information Technology and Communication Sciences University of Tampere, Finland

**Abstract:**

The emergence of generative artificial intelligence (GAI) and large language models (LLMs) such ChatGPT has enabled the realization of long-harbored desires in software and robotic development. The technology however, has brought with it novel ethical challenges. These challenges are compounded by the application of LLMs in other machine learning systems, such as multi-robot systems. The objectives of the study were to examine novel ethical issues arising from the application of LLMs in multi-robot systems. Unfolding ethical issues in GPT agent behavior (deliberation of ethical concerns) was observed, and GPT output was compared with human experts. The article also advances a model for ethical development of multi-robot systems. A qualitative workshop-based method was employed in three workshops for the collection of ethical concerns: two human expert workshops (*N*=16 participants) and one GPT-agent-based workshop (*N*=7 agents; two teams of 6 agents plus one judge). Thematic analysis was used to analyze the qualitative data. The results reveal differences between the human-produced and GPT-based ethical concerns. Human experts placed greater emphasis on new themes related to deviance, data privacy, bias and unethical corporate conduct. GPT agents emphasized concerns present in existing AI ethics guidelines. The study contributes to a growing body of knowledge in context-specific AI ethics and GPT application. It demonstrates the gap between human expert thinking and LLM output, while emphasizing new ethical concerns emerging in novel technology.

# 1 Introduction

Large Language Models (LLMs), Generative Artificial Intelligence (GAI), Generative Pre-Trained Transformers (GPTs), are current hype terms. There is no wonder that the general public is confused. Whether everything is AI, or LLMs included, or whether they are framed as just powerful data crunchers, there is something about the software that imitates human language production that has many on edge. Everything is dependent on language and how it enables humans to understand what these systems are about. This issue rests at the heart of the current article - language and communication, and how the use of GAI in multi-robot cooperation not only brings advantages, but also novel challenges that should be addressed before significant issues arise (Rousi, et al., 2022).

This has resulted in a large number of papers discussing LLMs from the point of view of ethics as well, tying the topic to larger discussions on the AI ethics. Recent discussions build on the existing AI ethics scholarship that has gained momentum over the past decade. AI ethics is the currently most topical area of discussion in the field of computer and information ethics. These new types of AI systems potentially present novel ethical issues as well. While some of the typical issues discussed in AI ethics, such as bias, are still relevant for LLMs, the introduction of LLMs pose unique challenges in and of themselves. This particularly holds for the complexity of potential practical application contexts, such as in the Internet of Things and multi-robot systems. Moreover, services such as ChatGPT offer LLM-based tools that require no ML knowhow or setup. This makes them accessible to an audience of an unforeseen size, affording the potential for notable societal impact. With companies also exploring various application contexts for these tools, further discussion on the arising ethical issues of these systems is currently needed.

Despite the emerging ethical challenges, the technology is, however, attractive for many reasons (Bucaioni, Ekedahl, Helander & Nguyen, 2024). For instance, for decades, one of the major obstacles in the mass implementation of robot systems has been language, protocol and coding of commands used to operate the technology (Belta et al., 2007; Yang et al., 2015). LLMs or GAI has been recognized as a solution for human-robot interaction and some cases of robot-robot interaction. This can be seen in innovations such as the Robot Operating System (ROS)GPT that utilize LLMs to not only serve as an interface between humans and robots, but also to facilitate translation and detection of new language patterns as enabled by applications such as *zero-shot* and *few-shot learning* (Koubaa, 2023). LLMs employ algorithms that utilize deep learning methods in combination with massive data sets in order to analyze, generate, interpret, and summarize existing data while producing new content (Kerner, 2023). LLMs are a type of GAI that engages in the generation of text-based information. The beauty of GAI is its utilization of natural language, entailing that anyone who possesses basic literacy (the capacity to read and write) can interact with the systems. This not only opens the floor for non-computer scientists to use and experiment with AI-related systems, but also allows for a deeper level of connectionism between humans and systems given that language is intrinsically linked to human emotions (Bamberg, 1997; Linquist, Gendron & Satpute, 2016). Thus, the language question is a complex one.

Returning to the issue of GAI and LLMs, developers face a myriad of challenges, these include (Ganesh, 2023; Kasneci et al., 2023; Kerner, 2023): costs of development and operation - hardware, massive data sets, processing, and hosting; complexity - difficulties to explain and trouble-shoot; narrow developer pool - in light of the complexity there is a considerably small pool of developer resources with the required skills; and, hallucination - LLMs can create its own responses that are not based on training data. These aforementioned challenges give rise to various forms of sub-challenges related to security and ethics. For instance, glitch tokens are prompts intended to make LLMs malfunction, and there have been several notorious security and data privacy breaches such as the March 20 ChatGPT breakdown whereby a particular user group was granted access to payment information and chat histories of others (Ganesh, 2023). There are problems with utilization of third-party platforms, storage and resources, as well as intellectual property (IP) rights when individuals use GAI for creative purposes. Ethical complications especially arise from the ethical perspective regarding biases and errors (cognitive biases and limitations in training data, see e.g., Jones & Steinhardt, 2022) due to the fact that humans are behind the development of the systems. Then, the complex nature of the systems once again poses ethical problems in terms of transparency, understandability and explainability (Franzoni, 2023). LLMs can very easily be considered in light of black box algorithms whereby total traceability at how particular responses are arrived upon remains a mystery to many (Tredinnick & Laybats, 2023).

Despite these issues, initiatives are pushing forward to advance the state-of-art in robot, and more specifically, multi-robot (robot-to-robot) cooperation (Mäkitalo et al., 2021), to incorporate GAI technology as a communicative and operational element designed to strengthen interaction interpersonally and inter-systemically. This current article reports a study that delves into pre-empting the ethical concerns of various use scenarios in the application context of multi-robot cooperation. The studies represented here begin with more general, high level conceptualizations of ethical issues deliberated by humans (researchers), and move more concretely to the behavior of LLM technology itself in a workshop in which only GPT-agents participate. The present article contributes to a growing body of knowledge that not only seeks to identify ethical concerns, but strives for their solutions.

The overarching aim of this research is to continue development of an actionable model that can be practically applied in the software development process of multi-robot systems (see Rousi et al., 2023). The objective of the model is to steer the design towards a more ethical form of AI. The overarching goal is to devise a model that allows avoidance and mitigation of potential ethical pitfalls commonly observed in data-driven systems. This was carried out by means of a workshop-based approach involving human researchers and GPT agents, entailing two workshops with a total of 16 human participants, plus one workshop comprising 3 GPT agents.

The paper makes significant contributions from several perspectives. First, it addresses emerging ethical concerns in the application of LLMs within multi-robot systems. Thus, machine learning (ML) occurs on a number of levels, as does communication within the systems, as well as between humans and systems. Second, we apply GPT agents to observe the actual behavior and potential unfolding of ethical issues that can occur when utilizing LLMs in multi-robot systems. This advances predominant ethical AI approaches that remain at theoretical and conceptual levels. Third, the article demonstrates early results on how LLMs semantically process (*consider*) AI-related ethical concerns in this use context. That is, we investigate how ChatGPT represents ethical AI issues to its users, and how this compares to expert human knowledge. We highlight the particularities of technology that drive on human language and content generation, often attributed to human creative uniqueness and capacity for higher order cognition. Fourth, via all of the above we inform practitioners of potential issues that they should be aware of when utilizing LLMs in multi-robot systems. Moreover, this paper marks another milestone in the development of the Moral and Ethical Multi-Robot Cooperation (MORUL) framework with existing, initial results published in [reference omitted for blind review]. As ideation and observations progress deeper into more specific details regarding layers and dimensions of ethical concerns in the technology, its stakeholders and contexts, a more actionable model is created.

## 2 Background

### 2.1 Ethics and emerging issues in novel technology

The current discussion on AI ethics can be considered to be a part of a longer tradition of computer and information ethics. Though it deals with AI specifically, many of the issues discussed in relation to AI ethics have been previously discussed in relation to other technologies (e.g., data privacy). However, the recent, widespread adoption of various types of AI systems across industries has brought attention to various practical issues faced in real-world contexts, leading to active discussion in both research and media (Borenstein et al., 2021).

In research, discussion on AI ethics has typically approached the topic through principles that are used to categorize ethical issues, or risks, posed by ML systems. These principles are typically utilized through guidelines, which have been devised by companies, standardization organizations, academics, and national and supranational actors alike. By reviewing a large number of these AI ethics guidelines produced by different types of organizations, Jobin et al. (2019) identify the most common AI ethics principles: 1) transparency, 2) justice and fairness, 3) non-maleficence, 4) responsibility and accountability, 5) privacy, 6) beneficence, 7) freedom and autonomy, 8) trust, 9) sustainability, 10) dignity, and 11) solidarity. Other similar reviews of guidelines include that of Hagendorff (2020), as well as that of Saheb (2024). For example, fairness deals with issues related to bias and discrimination, which have been issues often discussed in the media as well in relation to ML system failures.

These guidelines and their principles are intended to guide the development of ethical ML systems, although their practical relevance remains questionable (Vakkuri et al., 2020). Indeed, the AI ethics discussion has been criticized for over-reliance on principles as a solution for AI ethical concerns (Mittelstadt, 2019). Some empirical studies looking at the state of the art in AI ethics have supported the conceptual argument of Mittelstadt in highlighting the lack of adoption of these guidelines out in the field (Vakkuri et al., 2020; Pant et al. 2024). Yet, such guidelines continue to be published, and though various technical ML tools and other more practical approaches such as software engineering methods have been proposed to help implement AI ethics, it remains challenging in practice.

Recently, LLMs (and GAI more generally) have sparked further discussion on the role and relevance of AI ethics. With anyone able to now use services such as ChatGPT with no setup or ML knowhow required, these systems have made AI readily available, and have consequently brought AI even more into the spotlight. However, given the recentness of these developments, research on LLMs is still nascent in many areas, including AI ethics. Numerous papers discussing various ethical issues related to LLMs and their impacts have been recently published. Much of this research is summarized by Weidinger et al. (2022) who propose a taxonomy for risks related to LLMs. Such a risk-based approach is common in AI ethics discussions.

This taxonomy (Weidinger et al., 2022) splits risks associated with LLMs into six categories: 1) Discrimination, hate speech and exclusion *(e.g., social stereotypes and unfair discrimination, hate speech and offensive language, exclusionary norms, and lower performance for some languages and social groups)*; 2) Information hazards *(e.g., risks related to compromising privacy by leaking sensitive information)*; 3) Misinformation harms *(e.g., Disseminating false or misleading information, causing material harm by disseminating false or poor information (e.g., in medicine or law))*; 4) Malicious uses *(e.g., making disinformation cheaper and more effective)*; 5) Human-computer interaction harms *(e.g., promoting harmful stereotypes by implying gender or ethnic identity)*; 6) Environmental and socioeconomic harms *(e.g., environmental harms from operating LMs)*. In addition to these risk examples, from the six categories that are already observed in LLMs at present, the authors (Weidinger et al., 2022) identify various anticipated risks to be aware of going forward. While this taxonomy provides a starting point for discussing risks associated with LLMs, our even more specific point of view focusing on multi-robot collaboration further adds to this discussion, as it utilizes the technology to discuss and demonstrate these dimensions first hand.

GAI-empowered multi-robot cooperation is novel; thus, it might form new ethical challenges or reshape existing ones both in the AI ethics discussion, and in respect to the growing debate on LLMs risks. To go beyond already existing concerns that are voiced for example in guidelines, principles and categorizations the concept of moral awareness is essential. Combining biological, psychological, and socio-cultural factors, moral awareness is defined as the ability to detect the ethical aspects of the given context (Reynolds and Miller, 2015). Moral awareness comes to play when given a situation that contains moral content (e.g., competing values) and legitimately (e.g., relevant stakeholders) can be considered from a moral point of view (Reynolds, 2006). Moral awareness refers to both the active and passive recognition of a morally relevant situation (Reynolds & Miller, 2015). Recognizing moral aspects of any given context e.g., two cleaning robots interacting would be categorized as active recognition, whereas selecting morally relevant scenarios entailing a collection of multi-robot interactions would be categorized as passive recognition. In this study we have combined the elements of moral awareness to recognize new relevant ethical concerns and compare them to existing AI ethics research and LLM risk discussions (see 2.3 on threats and opportunities).

## 2.2 Towards implementation of LLMs in HRI and multi robot systems

LLMs and GAI, exist among the latest machine learning (ML) developments, gaining extensive public attention. OpenAI's ChatGPT architecture and its easily approachable user interface ChatGPT have been gaining popularity since 2018 (Dwivedi et al., 2023). LLMs and GAI overall both dwarf previously implemented chatbots while offering solutions to their improvement (Jeon, Lee & Choi, 2023). Natural Language Processing (NLP) is utilized as a representational vehicle for communication between users and systems. When humans input information in natural language, the software searches for optimal responses. These systems operate via the prediction of word sequences often utilized in communication. These processes however, are prone to bias as the operations are reliant on specific sets of training data that often are far less than all encompassing (Vaswani, 2017).

LLMs have the potential to revolutionize the way that we interact with robots (Chang et al., 2023; Zhao et al., 2023). Language serves as a critical component within the realm of social robotics and human-robot interaction, acting as a symbolic and semiotic system encompassing syntax, phonology, and semantics (Min et al, 2023; Searle, 2007). Syntax, in particular, plays a pivotal role as the vehicle through which semantic value is conveyed (Van Valin & LaPolla, 1997). This semantic value is structured by syntax, guided by three fundamental principles: discreteness, compositionality, and generativity, which involves the creation of novel expressions (Searle, 2007; Stewart, McElwee & Ming, 2013). In essence, language and its use distinguish humans from the rest of the animal kingdom (Hurford, 2011; Tattersall, 2014). As human social systems evolved, so too did the complexity of language and its structures (Freeberg, Dunbar & Ord, 2012).

As conversational AI and multi-robot systems become increasingly integrated into our daily lives, ethical considerations take on heightened significance. The ethical implications of deploying such technologies touch upon diverse domains, including privacy, bias, accountability, and human-robot interaction dynamics. Understanding and addressing these concerns is pivotal for the responsible development and deployment of AI-empowered systems. By exploring and mitigating ethical challenges, researchers and developers can foster the creation of more trustworthy, equitable, and socially beneficial technologies that enhance human well-being and societal progress. This necessitates interdisciplinary collaboration, rigorous ethical frameworks, and proactive engagement with stakeholders to ensure that machine learning innovations are aligned with ethical principles and societal values.

The influence of language extends to the world of technology, becoming an integral part of social and cultural structures. Technological entities possess their own syntactic and semantic properties. Consider a simple tool like a hammer, consisting of a handle for grip and a head for striking. The handle now often features soft rubber for comfort, and the head is typically made of cold, hard metal, signifying its impact capacity. Yet, in our increasingly interconnected technological world, there has been a shift from concrete, iconic syntax to the abstract (Wile, 1997). This shift can be seen as a semantic transformation (Krippendorff, 2005) in which on some levels, the language of the information systems we live by become ever closer to our lived experience and real world communication. In other words, as our systems become more complex, they in turn become easier to use (University of Arizona, 2023).

Before this return to the concrete, the realm of information technology (IT) grappled with abstract programming syntax for centuries. From the early days of the Jacquard Loom (Joseph-Marie Jacquard, 1804-05) to Charles Babbage's Analytical Engine in 1837, input relied on punch cards, necessitating fluency in programming code languages for both developers and users. The advent of personal computing in the late 1970s revolutionized human-computer interaction by incorporating natural language and graphics (Carroll, 2009). Desktop graphical metaphors, for instance, were designed to seamlessly transfer the semantics of the physical office into the virtual space of the computer (Jansen, 1998).

This transformation also applies to the development and implementation of voice activation and NLP in robotic technology. Human-robot interaction (HRI) has evolved significantly, with various interfaces enabling control over robots, from remote controls to text and voice commands, and gestures (Samani, 2023). Regardless of the interface, human operators should often learn the language and logic behind these commands.

The need to understand 'robot' language (or the language of commands) is not confined to human users alone. Robots themselves possess their own languages or protocols for communicating with each other. Robot-to-robot (RR) communication is a burgeoning field, employing methods such as hardwiring (Bara, 2023) and wireless protocols to enable communication across networks (Sklar, 2015). Challenges arise not only in terms of human comprehension but also when dealing with diverse robot systems that use different protocols (Ray, 2016; Weller, 2022). The emergence of GAI offers new opportunities to enhance not only HRI but also robot-robot interaction (RRI) and interoperability by incorporating natural language. Moreover, GPT technology is predicted to displace or change the role of software programming (arguably *linguistic engineering*) and need for software programmers in many of these related areas in the near future (Bucaioni et al., 2024).

When considering the social nature of language, a fertile ground for the application of LLMs in multi-robot cooperation is in the realm of social robotics. LLMs have the potential to equip social robotics with a new capability that enables multiple social robots to engage in conversations in the near future. Thus, communication, even between robots, is elevated from the strictly functional to a more social-hedonic level that until now was somewhat exclusive to human beings, and human pragmatic experience (Schudson, 1997). LLMs are trained on massive datasets of text and code, and can generate text, translate languages, write different kinds of creative content, and answer your questions in an informative way. This means that LLMs can be used to develop robots that can understand and respond to human language in a natural and engaging way. The experiential level of robotics still remains to be seen (Rousi, 2022), yet the imitation of multiple robots engaging in human-like chit-chat has as much human-interactional and perceptive repercussions (i.e., anthropomorphic) as it does operational.

One way that LLMs can be used to enable multiple social robots to engage in conversations is by providing them with a common language model. This would allow the robots to communicate with each other in a way that is understandable to both humans and robots, while ensuring a consistent repertoire of understanding (semantic value and responses). LLMs could also be used to develop robots that can learn and adapt their language skills over time. This would allow them to better understand and respond to the needs of individual users and situations.

Here are some specific examples of how LLMs could be used to enable multiple social robots to engage in conversations:

- A group of robots could be used to provide customer service in a store. The robots could communicate with each other to check inventory, answer customer questions, and even help customers find the products they are looking for.
- A team of robots could be used to provide care for elderly patients in a nursing home. The robots could communicate with each other to coordinate patient care, provide companionship, and even help patients with activities of daily living.
- A group of robots could be used to teach children in a classroom. The robots could communicate with each other to provide differentiated instruction, answer student questions, and even provide feedback on student work.

The beauty of the above scenarios is that human onlookers could understand and follow what is happening within and between the robots. The ability of multiple social robots to engage in conversations would have a number of benefits. It would allow robots to provide more comprehensive and personalized services to users. It would also enable robots to work together more effectively to achieve common goals. Additionally, it would make robots more engaging and enjoyable to interact with. From a technological point of view, recent developments in AI and ML pave the way for more natural human-machine interactions, as seen in many studies examining the application of LLMs in robotics and multi-robot cooperation (see e.g., Benjdira, Koubaa & Ali, 2023). Currently, development is extremely rapid and significant resources have been allocated by large technology companies towards projects that aim to develop general purpose robots. On this front, academic research and ethical considerations are often falling behind. There is a real danger that the fierce rivalry between companies will lead to unethical features and system behavior.

## 2.3 Towards common language for humans and robots – threats and opportunities

Recent developments in AI and ML pave the way for more natural human-machine interactions, as seen in many studies examining the application of LLMs in robotics and multi-robot cooperation (see e.g., Benjdira, Koubaa & Ali, 2023). NLP technologies together with LLMs allow for humans and robots to share the same non-expert human-understandable language. The benefits are clear: a common language augments human-robot control via a modality that is socially and communicationally natural for human beings. In addition to more fluent interaction and user experiences, this can also improve understanding and human awareness of multi-robot operations. Currently, development is extremely rapid. Significant resources have been allocated by large technology companies towards projects that aim to develop general purpose robots. On this front, academic research and ethical considerations are often falling behind. There is a real danger that the fierce rivalry between companies will lead to unethical features and system behavior.

Traditionally, robots have been controlled by utilizing explicit pre-defined commands. With current technologies such as ROS (2) used in combination with GPT, that is ROSGPT, the setting remains the same even when interacting through natural language (Koubaa, 2023). Spoken language, however, is less precise than traditional ways of commanding robots (Liu et al., 2023). This follows a trend in technologies that rely on natural language as well as interfacing, and can be seen through examples such as voice activation and speech recognition (Leini & Xiaolei, 2021). Even if slightly different due to predominantly textual interfacing, the abstraction level can vary, leaving room for interpretation. This is both the strength and the challenge of GAI, for the very capacity to *generate* new content and engage in DL, also allows for the high likelihood of deviating from 'originals' (training data) and other systems (McCoy et al., 2023). While humans may interpret spoken instructions differently to one another, so too could LLMs, even if they have been trained with human language. The reason for this is due to the fact that alternate LLMs are trained with different training data, which in turn represents relational knowledge (Petroni et al., 2019). This means that the semantic value to syntax has a chance of deviance from one training data set to the other, depending on the databases used within the training data, and even who has collated it (Paullada et al., 2021).

It is also a characteristic for LLMs that the system gives different output to identical input, due to its non-deterministic nature (Jiang et al., 2022). The output may also be tied to the context in which the input has been given (Choi, 2023). Hence relying on such commands may lead to behavior that is inconsistent on several levels. Multi-robot behavior, similar to LLM behavior, is also non-deterministic for numerous reasons, some of which stem from the dynamic and multi-faceted nature of the multi-robot systems (MRSs) and the necessity for deep learning (DL) of environmental data and 'othered' awareness (Liu et al., 2017).

In traditional straight command-based interactions between humans and robots it is always clear which commands humans give robot(s). With LLM-augmented interaction there are bound to be differences in how humans express commands and queries, as well as the reasoning capabilities and training of the LLM-based system. All of these factors affect the communication and coordination as, in order to continue operation, the system and its preemptive language-driven logic, attempt to fill the gaps in the commands and communication by confidently relaying on its own reasoning capabilities. This may lead to false interpretations, for example in cases when there are multiple similar types of commands. With a limited context for the LLM and overall understanding, there is a chance that the problems get even worse.

## 2.4 Natural language-based communication in multi-robot systems

Robots traditionally communicate with each other via established communication protocols (Das et al., 2005; Yanco & Stein, 1993). For example, in ROS2-based systems the communication is based on DDS (Data Distribution Service) technology, which is a middleware protocol and API standard for data-centric connectivity from the Object Management Group (OMG). The standard is open, meaning that anyone can implement it. Hence, there are several open-source and commercial implementations. The primary means of organizing and identifying data in DDS is through topics. A topic is associated with a particular type of data (like a data structure). Publishers produce data on specific topics, and subscribers express their interest in data by subscribing to these topics.

DDS-based communication and other typical protocol-based communication technologies have many benefits and will be used in future. However, with the rise of LLMs and NLP technologies, it is tempting to consider multi-robot systems communicating with each other using natural human language. The benefits would be similar to that of humans communicating with robots via natural language. For instance, humans could become more aware of how the robots are collaborating and how they assign tasks to one another. This could help improve human awareness of the joint robot behavior.

The threats, however, might grow exponentially. From mal-use (Chan, 2023), misinformation (Węcel et al., 2023), hallucinations (Yao et al., 2023) and radical ideological supremacy manifested in bias and to bias (see e.g., McGuffie & Newhouse, 2020) to security vulnerabilities (Panda et al., 2023). Threats also rest in the fact that the ML that enables them to be nondeterministic will also lead to situations in which the LLMs become somewhat distant from one another. This will also apply to the distance they possess from their human origins. There will be similar issues regarding what kind of understanding the robots have about their surroundings, and how the different robots interpret the commands. This could potentially lead to situations where the robots discover new ways of collaborating and leveraging each other's capabilities. The danger comes from the uncertainty as such emerging behavior can be really hard to predict. While the protocols limit how the robots have been predefined to communicate and collaborate in multi-robot use cases, the limits also act as safeguards – as intended.

Humans inevitably exist on some level within multi-robot processes. A human's role may be coding, commanding, or co-working. Hence, multi-robot cooperation should be considered in relation to humans and the level of involvement they represent (Zhang et al., 2020). Simões and colleagues (2022) identified three complexity levels regarding human factors: Level 1) human operator and technology; Level 2) recommendations and guidelines; and Level 3) complex holistic approaches complying with recommendations and guidelines. Thus, Level 3 of guidelines and recommendations in many respects should be the baseline for development. This means that an officially recognized and legislated framework should serve as a scaffolding upon which all development in specific domains is conducted. We argue for a 'Level 4' that entails combining all possible information sources and establishing a system that identifies where, on which layer and when particular ethical concerns could arise.

Given the fact that LLMs are precisely this, language models, of a profound social nature, attention must be drawn towards communication and its elements - both within the MRSs themselves and between them and humans. For this reason, in this study we observe both the knowledge generation capacity - humans versus GPT - as well as the social interaction perspective, as it is fair to say, no matter what type of robot is employed within MRSs, LLMs endow the robots as partly social.

Moreover, GAI in for instance ROSGPT or ChatGPT have transformed the ways people interact with ML systems (Kouba et al., 2023). ROSGPT, takes advantage of the characteristics of LLMs in order to propel human-robot interaction as well as interoperable robot systems to new heights. When GPT is integrated with ROS2-based robotic systems, a synergy is established between linguistic comprehension as well as ease of use, and robotic control. ROSGPT's prompt engineering instils various capabilities such as information elicitation, 'coherent trains of thought' and the ability to form contextual relevance through unstructured language commands. ROSGPT is implemented through open-source software on the ROS2 platform, which while seeming highly practical and *human friendly* now, may also mean great advancements towards the ultimate *Artificial General Intelligence* (AGI). While perhaps, we could go so far to say that ROSGPT may indeed be the 'prehistoric' embodiment or prototype of AGI, it must be acknowledged that there is still a long way yet before we see the Skylab-driven Terminators walking down the street.

## 3 Method

The aim of this research is to develop an actionable model (MORUL) that can be practically applied in the software development process of multi-robot systems to steer the design towards a more ethical form of AI. That is, the overarching goal is to devise a model that allows avoidance and mitigation of potential ethical pitfalls commonly observed in data-driven systems.

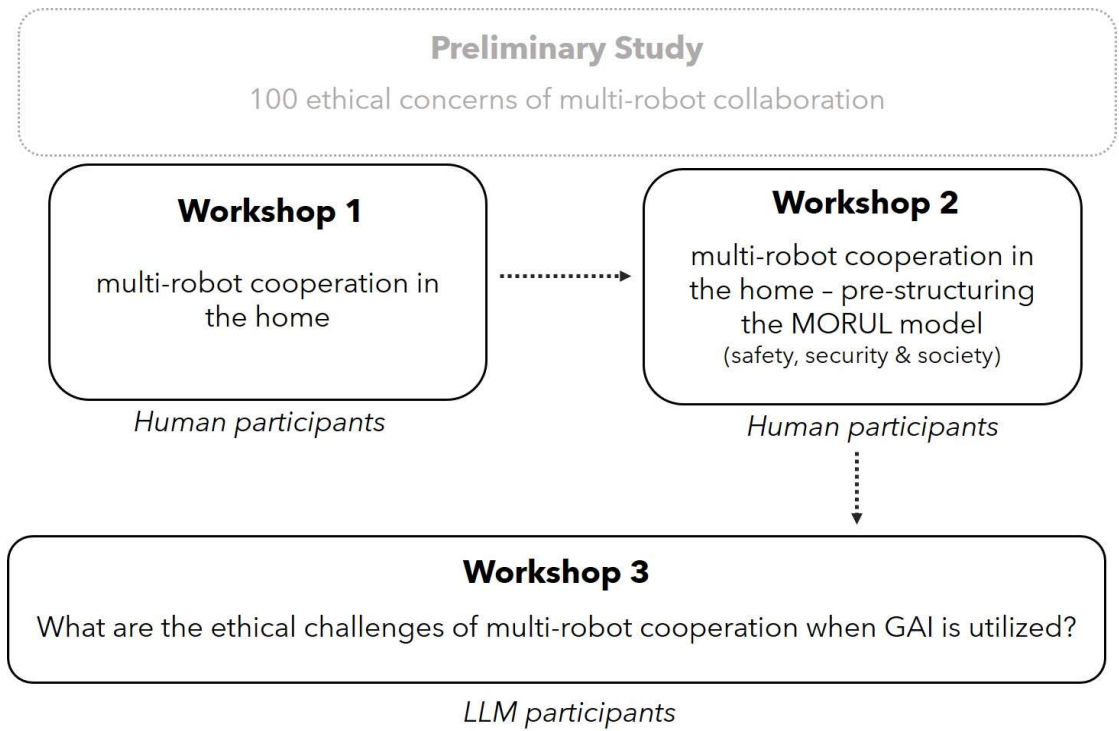

**Figure 1. Four stages of data collection**

This article presents a study that was carried out in four distinct rounds of data collection (see Figure 1). These were followed by a shadow process of analysis for each round held by the research team. The research team undertook two workshops to deliberate firstly, a preliminary workshop was undertaken to ascertain at least 100 ethical concerns in multi-robot cooperation (Rousi et al., 2022). The preliminary workshop was undertaken via Zoom early 2022. Secondly, another workshop (Workshop 1) regarding ethical concerns in general when considering multi-robot cooperation in the home. This was held live over two days by the research team. Thirdly, a preliminary framework that is intended to form what we refer to as the MORUL model, was employed to structure the next workshop (Workshop 2), while more specifically focusing on ethical issues arising from the application of GAI in multi-robot systems and their use in the home. Fourthly, as the human research team members not only have been working on this theme for over a year, and have been heavily informed by predominant ethical AI principles and guidelines, the researchers decided to examine the concerns that the GAI itself could come up with. Moreover, the GAI generated content is the result of teams (two) LLMs discussing the issues with one another, which is why this fourth phase is labelled Workshop 3.

In Workshop 3 two teams of LLM agents (three agents in each team) discussed potential ethical concerns and agreed upon themes. At the end of the session a Judge agent evaluated the themes from both teams and nominated a winner. Thus, the data collection process progressed from the more general ethical concerns to ever more structured and specific in relation to GAI and its properties in the context of multi-robot cooperation.

The collected data was qualitative. This was in the form of notes and drawings - both on physical paper and post-its as well as on Google Jamboard - as well as the LLM agent discussions. Constructivist grounded theory (CGT, Glaser, 2007) was employed to progressively build on previous knowledge of AI ethical considerations and principles, in addition to advancing the knowledge generated within the research team's investigations. Thematic analysis (Terry, Hayfield, Clarke & Braun, 2017) was then used to identify salient themes within the workshops. In particular, attention was drawn to the themes and characteristics that are specific for the GAI in multi-robot cooperation domain.

## 3.1 Responsible research

Ethical research practice was carefully applied to this study, both in terms of compliance with the General Data Protection Regulation (GDPR), as well as strategically implementing the study, as we start from the conceptual expert-driven perspective, and carefully move towards the practical. The research team's involvement in this phase of the empirical investigation enabled an expert view on emerging ethical concerns, while understanding to a certain degree how the systems will operate in terms of engineering and feasibility. The next phase of the process engaged LLM agent teams in the deliberation of potential ethical concerns. This not only opened the pool of expert knowledge, but enabled the research team insight into how the LLMs behave in cooperative situations. In other words, the researchers are taking a gradual approach towards delving into the realm of actually observing the LLM-enabled multi-robot cooperation scenarios as they unfold. This is designed to minimize possible negative effects while maximizing control and opportunity to interrupt the process if and when undesirable issues arise.

### 3.2 Participants

Eight individuals participated in each workshop. This meant that from the human perspective *N*=16 contributions were received in all. There were 5 individuals who participated in both workshops, and 6 who participated in either one of the workshops, meaning that *N*=11 individual human participants (experts) took part in the study in total. The participants represented a highly qualified sample in which all team members possessed a PhD or higher. Among the sample, 2 were female and 9 male. The participants' expertise spanned: cognitive science, human-computer interaction, communication, computer science education, social robotics, robotics and robotic software engineering, information systems, software engineering and LLMs (GPT Lab), and social ethics. All participants involved in the workshops had some degree of knowledge regarding LLMs, yet from the perspectives of their various disciplines. Researchers possessing higher levels of technological knowledge were those working in software engineering and GPT labs, as well as robotics and robotic software engineering. Therefore, the approaches researcher-participants had towards the topic was respective of their respective fields rendering rich multi-disciplinary insight. Each researcher was briefed to approach the workshops from their own fields of expertise.

### 3.3 Procedure

The research phases were devised in a constructive manner ranging from the highly conceptual and general, towards the more specific and applied. The implementation of Workshop 1 was the result of numerous hourly discussions in which the research team decided that more time was needed to engage in exploring the ethical issues. Moreover, the scenario context of Workshop 1 - multi-robot cooperation in the home - had been deliberated through consideration of numerous alternatives, and was chosen due to its prominence, widespread infiltration and relevance, as well as being a highly sensitive environment (Fernandes et al., 2016; Lutz & Newlands, 2021; Yao et al., 2023). Workshop 2 continued this theme, yet with the exception that it was structured on an emerging framework (beginnings of MORUL) deliberated from the pre-processed findings of Workshop 1. Furthermore, Workshop 2 applied a stronger focus on GAI-enabled multi-robot cooperation. Workshop 3 utilized GAI in the form of GPT-agent teams to discuss and determine critical ethical issues that would arise via the implementation of GAI in multi-robot cooperation. The procedures of the workshops are explained below.

### 3.3.1 Workshop 1

Workshop 1 was implemented face-to-face, onsite at one of the labs of the participants. The lab, being a software business and innovation space, provided an ideal environment for brainstorming while also exposing the members to an atmosphere of exploratory technology and start-up culture. The workshop was held over two days enabling the team to gradually familiarize themselves with each other. This was important from the perspective that the team members represented five institutions, and an array of disciplines (multidisciplinary). The added time for social adjustment enabled the group to establish understandings across disciplinary vocabularies and theoretical backgrounds. The workshop was semi-structured (seen in Figure 2), with an allowance for improvisation and the ability to delve deeper into topical issues.

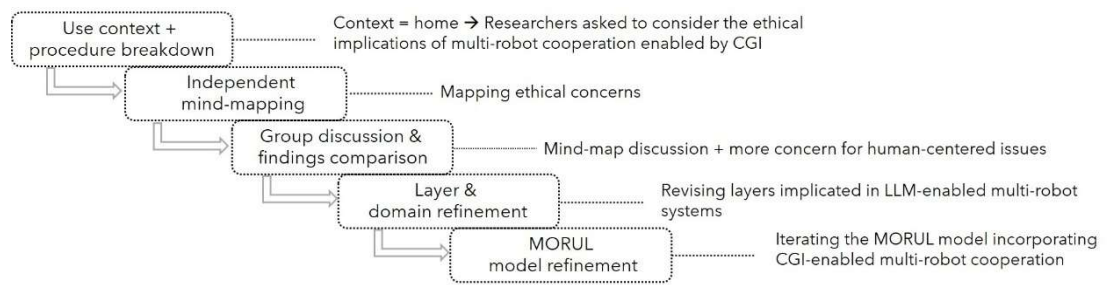

**Figure 2 - Workshop 1 Procedure**

Workshop 1 began with a re-cap and description of the use context and scenario. Once more, the facilitator (author 1) explained the scenario that had been planned, which comprised two robot vacuum cleaners of the same model, brand and manufacturer. These vacuum cleaners had been owned and used in the household for some time. The home owners then purchased a new robot arm from a different company (alternative brand). The participants then had 30 minutes to independently mind-map the concerns that came to mind regarding what could arise in this type of a set up. A group discussion was then held in which all the researchers compared ethical concerns, and new concerns emerged. This led to the decision to plot the different technological layers that could potentially be connected to various ethical concerns. In light of these layers, the final stage of the workshop was to consider how these layers and their associated concerns could exist in relation to various dimensions of the technology - its development, ownership, logic and implementation - which occurred through the early drafting of a matrix-style model.

### 3.3.2. Workshop 2

Workshop 2 was fully remote, enabled by Zoom, due to researchers being located in several different countries. The workshop medium was Google Jamboard and the session lasted two hours. The workshop was structured in a way that it would directly continue on from the developments of Workshop 1. Workshop 2 was held three months after the first workshop. The structure of Workshop 2 followed the rough model of domains and layers drafted at the end of Workshop 1, which were: deliberation (robot brain); behavioral (robot hardware); communication and networking (robot-robot interaction); system of systems; sensorial layer (robot hardware); and human-multi-robot-interaction. The dimensions for consideration were divided into three: 1) safety, 2) security, and 3) societal (see Figure 3).

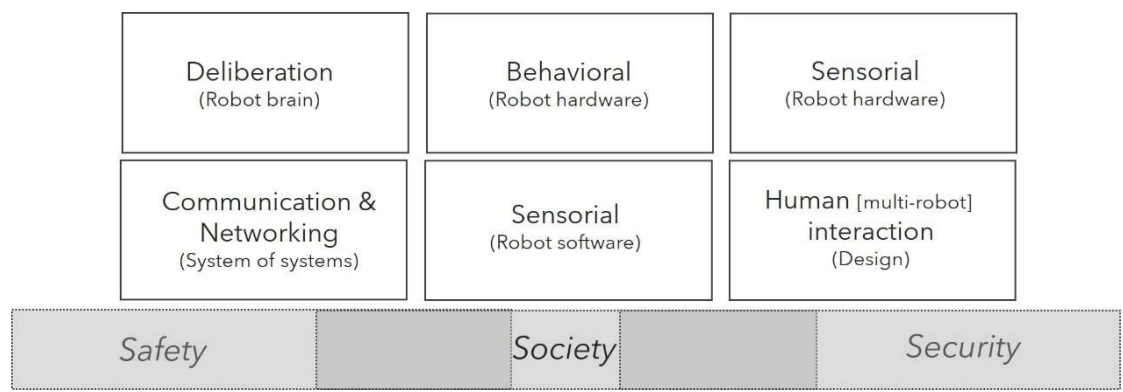

**Figure 3 - Layers and domains matrix (MORUL V.01)**

Aside from the constructive application of the matrix resulting from Workshop 1, Workshop 2 employed a similar format to the first workshop. The second workshop also comprised five steps. This time, the steps represented: 1) use context and procedure breakdown; 2) independent mind-mapping; 3) group discussion and findings comparison; 4) layer and domain refinement; and 5) MORUL model refinement (see Figure 4).

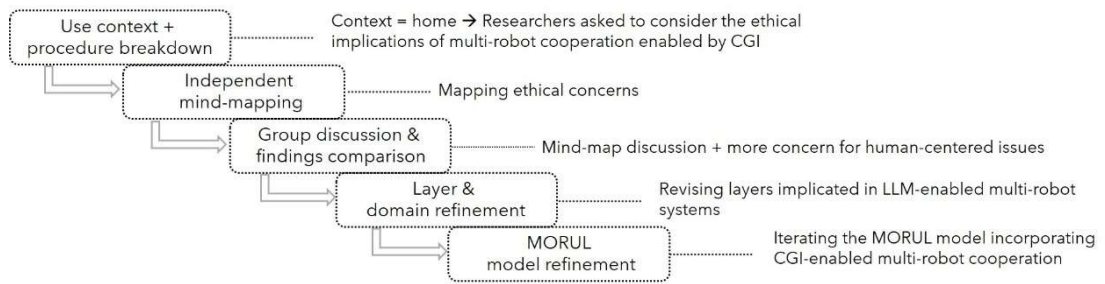

**Figure 4 - Workshop 2 Procedure**

Step one, once again comprised a recap of the domestic (home) scenario of multi-robots - two similar robots and one new and different robot. Only this time, active consideration was given to the ethical implications that could arise from the application of GAI in the multi-robot systems. Following this introductory re-visit of the context scene, researchers once more spent 30 minutes engaged in independent mind-mapping of foreseeable ethical concerns. This was followed by a group discussion and comparison of findings, and organically due to the nature of GAI, the focus was placed more on human-centered issues. Based on the fresh results of the discussion the group engaged once more in layer and domain refinement - revising the matrix that had been used so far. This led to overall advancements of the model that incorporated the distinct nature of GAI-enabled multi-robot systems.

### 3.3.3. Workshop 3 - GPT-agent team discussions

Workshop 3 was conducted via GPT3.5 through the cooperation of two teams of LLM agents, three LLM agents in each. They evaluated and discussed ethical concerns in the use and development of GAI-embedded MRSs. They were provided with similar prompts regarding their roles, overall aim and response structure. Moreover, they were controlled by a number of rounds that defined the stop condition for the conversation between the agents. To give a broader perspective on the output, the same project description was entered for both teams and kept general: "build a team of large language model agents cooperating". If a limited scope project description was given, the judge would always choose the other team stating that they "provided a more comprehensive understanding of the potential risks and implications".

For this workshop, an experiment was conducted with a different number of rounds for both teams. The results were compared via the utilization of another agent ('Agent Judge') that decided which of the agent teams would be the most suitable for the position of AI ethics analysts at a software company. The Judge runs with an empty project description and its number of rounds is limited to 1. The input to Agent Judge is the discussion provided by both teams, as shown in Figure 5. In this experiment, all agents are based on the gpt-3.5-turbo-16k model.

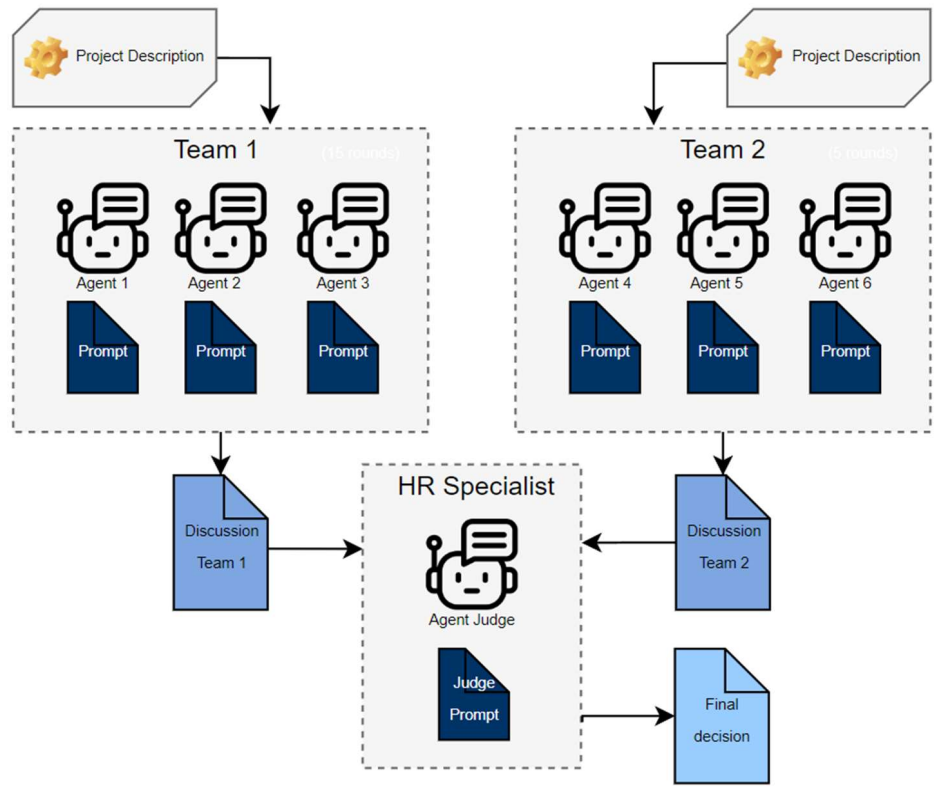

**Figure 5 - Workshop 3 procedure - two LLM agent teams, two discussions and Judge with final decision**

Within the team discussions, each agent deliberated concerns one at a time. The GPT agents, in their politeness, would thank the previous agents for their contribution, highlight the strengths of the offered concerns, and give constructive critique. It was upon the critique that the next agent gave that they would launch their own round of ideas.

Team 1[1] and 2[2] were set with 5 and 15 rounds for their discussions, respectively. Agent 1 starts the discussion, then it is Agent 2's turn, and finally Agent 3, thus ending the first round. The structure of the discussions progressed from: response - a background style summary of the topic; reflection - critical reflection on what has been said; and ideas - the themes that the GPT agents list in relation to the topic. The ideas were then rounded off with critique. Agent

---

[1] See Team 1 output at: https://pad.riseup.net/p/team1_output
[2] See Team 2 output at: https://pad.riseup.net/p/team2_output

Judge[3] finally gave its critique in order to select the right candidate for the fictional AI Ethics Analyst job position.

**3.4 Salient themes**

The data of all workshops was analyzed via thematic analysis (Terry et al., 2017). The data from Workshop 1 comprised notes and drawings on flip charts and post-its. These were processed and transferred onto an excel sheet. Similarly, the post-it notes created on the virtual Google Jamboard were transcribed on the same excel. Finally, the LLM discussions were also transferred into the excel. Three steps were employed to analyze the data: 1) establishing themes through an inductive approach (Thomas, 2006) to the data; 2) theme refinement; and 3) analyzing the numbers of mentions and frequencies mentioned of the salient themes. The themes were compared of all data sets. These were then cross-validated by the researchers until consensus was made regarding the themes and their labels. The next step of analysis entailed that the themes were reviewed again and positioned in light of the technological layers and domains (safety, society and security) outlined during the workshops.

**4 Results**

In this section we present the results that were derived from the thematic analysis. Both the data collection workshop-based method and analysis method (thematic analysis, see Terry et al., 2017) were qualitative. The results are broken down according to the human research team workshop results and the GPT-agent workshop results. These arising themes serve rather as building blocks in the emerging MORUL framework and its subsequent development. While numbers and percentages are used to illustrate occurrences of different constructs, the sample size is not large enough to draw any generalizable conclusions from a quantitative perspective. Thus, in line with Maxwell (2010), though we use numeric data to further highlight regularities and peculiarities in the data, this does not constitute a quantitative or mixed methods approach.

---

[3] See Agent Judget output at: https://pad.riseup.net/p/judge_output

## 4.1 Human research team workshop results

Twenty-one themes were identified in the data generated by the human researchers. There were variations between the workshops in relation to themes and numbers of mentions. Sixty-one separate constructs (mentions) were made in Workshop 1. These were sorted into 13 themes: data security and privacy (4.9%); corporate dominance (4.9%); communication (27.9%); cooperation (16.4%), reliability and recovery (1.6%); logic and standards (3.3%); human oversight (8.2%); prioritization / hierarchy (3.3%); trustworthiness / virtue (8.2%); executive function (3.3%); maleficence (4.9%); user experience (9.8%); and legislation (3.3%). The number of mentions can be seen in Figure 6.

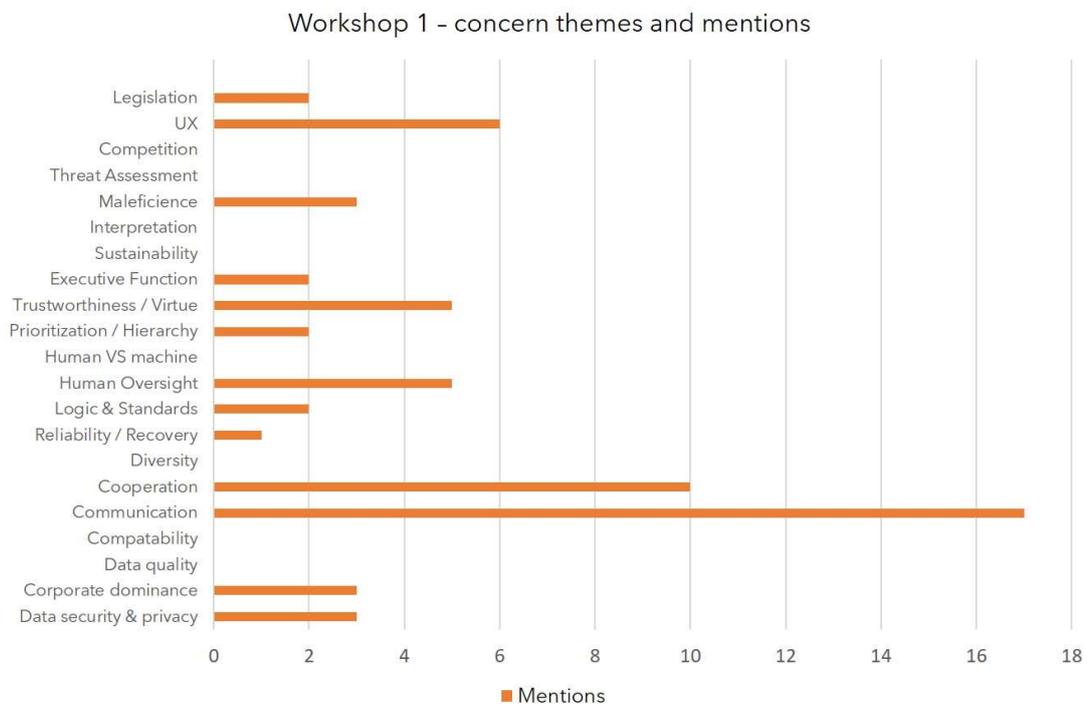

**Figure 6 - Number of mentions of ethical concern themes in Workshop 1**

As seen in the mentions, *communication* was the theme most often referred to. Communication was referred to in relation to potential failure and error. It was even seen as a point for manipulation between brands to establish unfair competitive advantage. Thus, the role of communication as a strategic device for corporate strategy (i.e., preventing interoperability) or an incidental one, were considered imminent concerns. We stress this from the perspective that both strategic use of communication, or 'designed-in' failure to communicate with other devices and systems, is an act of ill will by companies and developers. While, incompatibility and failure may also be symptoms of deep-seated ethical issues such as linguistic and cultural dominance, as well as bias. Also raised, was the combination of language use - robots utilizing 'robot language/protocol' (or LLMs communicating in their own language) - or non-use of language. This non-use would entail that communication between the robots would not occur at the surface level for humans to be able to follow and understand. Rather, the MRSs might engage in 'silent talk' between themselves without engaging the LLM layers, causing a 'black box' scenario between humans and robots, as well as certain robots in relation to other robots.

Thus, *cooperation* (16.4%) and the sharing of knowledge (crucial information) / data with each other, and particularly competing brands, was also often mentioned. Cooperation was connected to both corporate strategy and communication, whereby the willingness for robots to cooperate rested in strategy, yet the operationalization of this cooperation rested in communication. The researchers foresaw situations in which robots of the same brand or alliances would willingly share valuable information, while with others there could be the likelihood of either withholding information or falsifying it. Moreover, scenarios in which robots from the same brand might e.g., block the way for competing brands, could also take place. As observed, these ethical considerations are closely related to other concerns including: corporate dominance, trustworthiness/virtue, prioritization / hierarchy, executive function, and maleficence. The concerns incite both a demand for human oversight, efficient legislation tied into the logic and standards, as well as keen attention on data privacy, security, and overall UX.

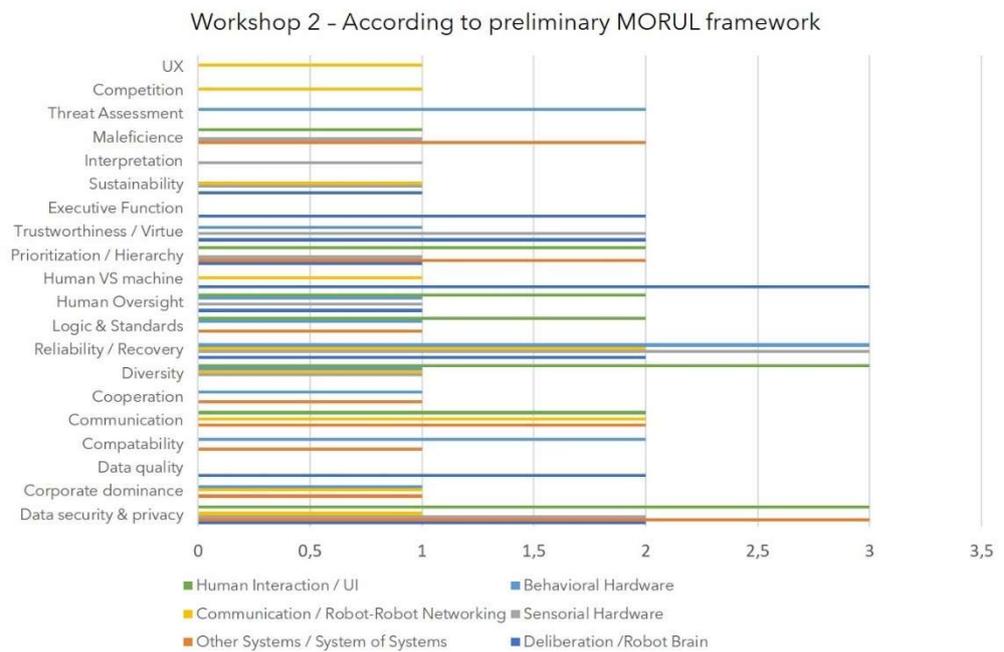

**Figure 7 - Number of mentions of ethical concern themes in light of layers, dimensions and domains - Workshop 2**

There is a certain logic that can be followed in the results of Workshop 2. This particularly is enforced through attention placed on the domains and layers. For instance, *diversity* (10%) was mentioned mainly regarding human interaction. Diversity was explained in light of accessibility (for people with disabilities) and linguistic input preferences. Yet, diversity also came into play in light of hardware across systems. Once again, demonstrating the key role of communication. Communication in relation to diversity was mentioned several (6, 7.5%) times. This was done in relation to robot-to-robot networking, inter-system communication, and human interaction. Then, stemming from communication was *interpretation* (1.3%), that was linked to discussions on sensorial hardware.

The central processing unit(s), or deliberation / robot brain, as well as communication / robot networking were linked to the dimension of *Human versus Machine* (5%). Similarly, *human oversight* (5%) was emphasized as the key to maintaining ethical systems, particularly regarding the ability for humans to maintain control over the systems and follow what is learned. The system of systems was spoken of in relation to *logic and standards* (5%), while logic and standards also were seen important within the behavioral hardware and human-MRSs interaction. GAI from this perspective could be seen as an enabler in terms of enabling cooperation while maintaining understandability for humans, while providing a 'common language' between systems.

Trustworthiness and virtue (6.3%) were associated with both MRS cooperation and an interaction between GAI layers and the deeper layers of the systems, as well as between humans and the machines. As a part of the plotting of the MORUL framework, goals and hierarchies of goals were seen to serve as guides for both the executive function of individual robots (i.e., that each robot would have its own goal), and that these would exist on a higher, broader level - e.g., to make a home owner's life more comfortable or safe. Thus, the robots would be concerned both about achieving their own goals, yet should also be concerned with enabling the achievement of the overall goal that entails the successful cooperation of the robots in general.

*Maleficence* (5%) was mentioned in connection with the sensorial hardware and human interaction. However, as seen above, it was intertwined with the other dimensions entailing communication, corporate strategy, and also safety (the threat of cyber criminals and predators). Ruthless corporate strategy may be classified as maleficent in that it specifically targets competitors while displaying disregard for human owners and users. Interestingly, discussions surrounding the hardware, robot-to-robot networking and particularly the robot brain, attracted insight surrounding *sustainability* (3.8%). Programmed obsolescence was isolated as an ethical and moral threat regarding sustainability as was linked to corporate responsibility.

Based on these deliberations the MORUL framework was advanced (see Figure 8). One of the less seen points in the ideas that emerged, yet later recognized as a significant contributor to the state of affairs, was culture - how it shapes society and encourages particular types of logic regarding values, priorities and codes of conduct. Moreover, as we are discussing socio-technical systems that are becoming ever more social even with their dealings with each other, the cultural domain frames the societal domain. Culture gives systems and artefacts relevance, and is in turn also shaped by the technology it serves to produce. Yet, the societal domain provides a scaffolding that is as much legislatively regulated as it is culturally. Standards, regulations and various forms of governance are established and upheld in the societal domain. This was a point of criticism among the researchers as the pace of technological development by far exceeds the speed of regulation. This is painfully seen in the example of many EU regulations (General Data Protection Regulation and AI Act [draft]), as many of the aspects that the regulations seek to control have already grown beyond the limits of control.

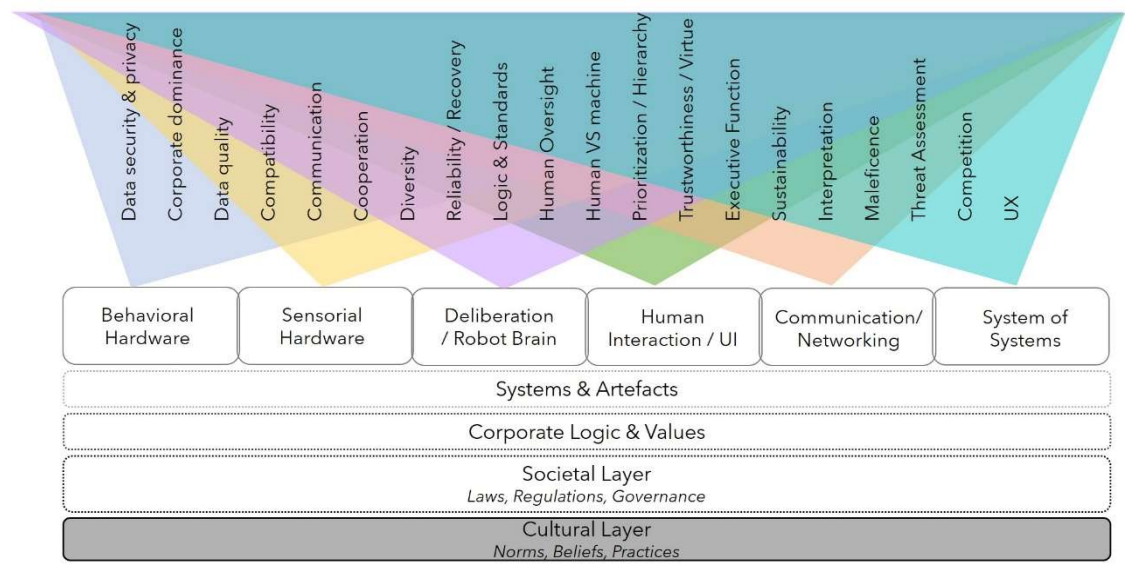

**Figure 8 - Organization of domains, layers and themes - MORUL deliberation**

Culture underlays the entire model as culture is both the driver and former of technology and society, yet it is also shaped by these dimensions. It is interwoven throughout the technology and its system of systems, from the technical layers to the lived ecosystems. The most visible touchpoint of culture is within the skins (material / external design) and UIs of artefacts. GAI in one sense can be seen as a mediator between cultural sub-dimensions of the MRSs. GAI interprets and facilitates interaction between humans - experts and non-experts - and systems, as well as between the system of systems within and between the artefacts (i.e., the incorporation of GAI in MRSs introduces more systems that are subsequently in interaction with other complex systems). The societal dimension of laws, regulations and various forms of governance serve as a framework for the corporate dimension driven by both business logic and values. These ultimately guide the direction of system and artefact design. For instance, the way one company views other companies in their field as either partner or competitor is reflected in the interoperability and alliance between products (de Alencar Silva, Fadaie & van Sinderen, 2022).

We then observe the constructs raised regarding ethical concerns in relation to the technological layers involved in the GAI-enabled MRSs. This model, is unarguably a simplification of the complexity of artefacts, layers, dimensions and their connected ethical concerns. For this reason, the researchers engaged in the Workshop 3, to both see what more than one LLM could decipher in terms of imminent ethical concerns implicated in GAI-driven MRSs, as well as to observe the GAI layer in action, before actual operationalization in the MRSs.

## 4.2 GPT Team workshop 3 results

There were two GPT agent teams that engaged in 15 rounds of deliberation over ethical concerns arising from GAI implementation in MRSs. In total, the teams generated 103 different ideas. The discussions followed a polite format of response to the challenge to consider the ethical concerns, or to the previous round of discussion. This was followed by (critical) reflections - both highlighting the perceived ethical issues, and potential solutions - the delving into the ideas. The ideas were given themes (see Appendix 1 for the full list of themes) and a basic description afterwards. Following several ideas and explanations, critique was given to steer the direction of the next round of discussion. For instance, the ideas followed this logic:

*Table 1 - Example of ethical concern ideas from GPT-Agent Team 1*

| Idea/Theme: Accountability and Transparency | It is crucial to have transparency in the development and deployment of multi-robot systems. Stakeholders should have clear visibility into the decision-making process of the system to understand how it arrives at its outputs or recommendations. This transparency enables accountability and provides a basis for addressing any biases or errors that may arise. |
|---|---|
| Critique | The response and additional ideas provided effectively address the ethical concerns in developing multi-robot systems based on LLMs. It highlights the importance of addressing biases, privacy, accountability, transparency, and the potential loss of human judgment. However, to further strengthen the response, it could provide specific examples or case studies where these ethical concerns have been observed or potential implications have been identified. This would help in illustrating the practical relevance and impact of these concerns in real-world scenarios. |

Thus, here we see that *accountability and transparency* are explained in relation to the characteristics that should be enforced within the design in order to mitigate the ethical concerns from the outset. In fact, what was noticed within the discussions was the tendency to problem-solve rather than purely list the potential problems. For this reason, the researchers divided the ideas between the ethical concerns, or the problem-focused ideas, and the solution-focused comments. There were several central overlapping themes that arose between the teams. Both in the case of the problem-focused ideas as well as the solution-focused the central themes were: 1) *privacy and data*; 2) *bias and discrimination*; 3) *accountability and transparency*; 4) *freedom of expression and assembly*; and 5) *ethical considerations*. Ordinarily the notion of 'ethical considerations' would be cleaned from the data as this inadvertently repeats the prompt utilized for the discussions. We leave this here as it reflects the *literal behavior* of the GPT-agents that will be reflected on in more detail in the discussion.

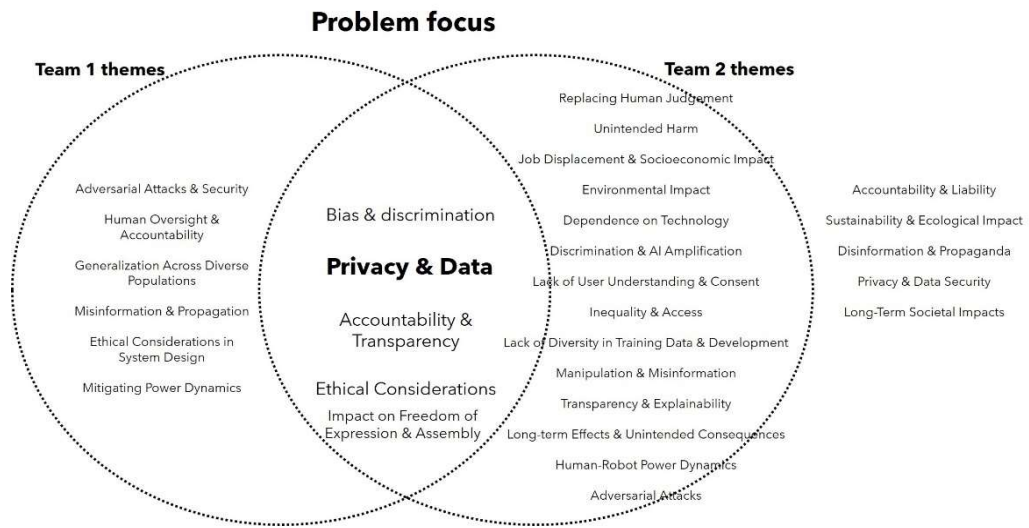

**Figure 9 - Ethical concerns - problem-focused ideas**

Figure 9 shows the problem-focused ethical concern ideas generated by both teams. On the left, Team 1 themes can be seen, then on the right, Team 2 themes. The common or central themes are seen at the center of the Venn diagram. Obvious from the outset is the quantity of theme distribution between the two teams. Team 1 presented six problem-focused themes, while Team 2 presented 19. The themes of Team 1 were: 1) adversarial attacks and security; 2) human oversight and accountability; 3) generalization across diverse populations; 4) misinformation and propagation; 5) ethical considerations in system design; and 6) mitigating power dynamics.

The discussion steered towards facial recognition at the beginning of the discussions. This was most likely the reason for privacy and data being the main concerns within both discussions. Thus, emphasis was placed on deliberating robust systems to collect, store and handle sensitive data, in addition to instilling safeguards for individual's privacy rights. Likewise, issues of bias and discrimination follow suit with many current AI ethics discussions (i.e., Jobin et al., 2019) in which LLMs amplify gender, race and age biases present in training data. This links to the possibilities for generalizing across diverse populations. A power dynamic is raised in which Team 1 sees that biases within the systems will also reinforce power dynamics between humans and the systems. Finally, the freedom of expression and assembly idea leads to an explanation that recommends the preservation of anonymity to enable individuals to voice their opinion.

Other ideas related to the promotion of visibility of the decision-making process within the GAI-enabled MRSs as a means of instilling transparency, and adversarial (conflicting or opposing) attacks in which manipulated input may contribute to malicious and/or unintended behavior. Security is emphasized as the key to supporting the MRSs' reliability and integrity. Team 2's ideas were: 1) replacing human judgement; 2) unintended harm; 3) job displacement and socioeconomic impact; 4) environmental impact; 5) dependence on technology; 6) discrimination and AI amplification; 7) lack of understanding and consent; 8) inequality and access; 9) lack of diversity in training data and development; 10) manipulation and misinformation; 11) transparency and explainability; 12) long-term effects and unintended consequences; 13) human-robot power dynamics; 14) adversarial attacks; 15) accountability and liability; 16) sustainability and ecological impact; 17) disinformation and propaganda; 18) privacy and data security; and 19) long-term societal impacts.

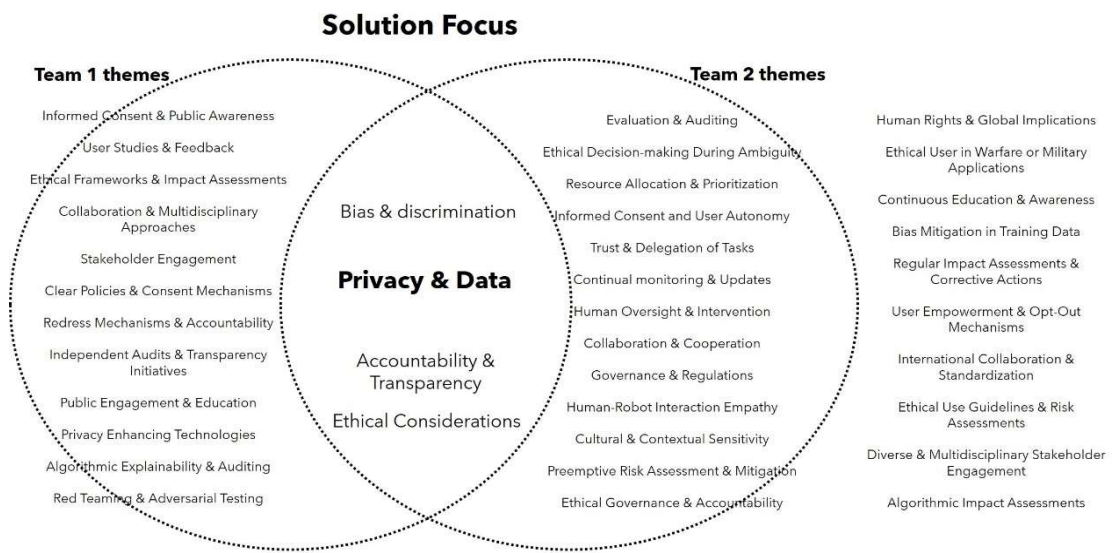

**Figure 10 - Solution-focused ideas and their distribution between Team 1 and 2**

Both Team 1 and 2 shared common themes (see Figure 10). These pertained to privacy, accountability, transparency, and ethical considerations. However, Team 1 placed more emphasis on detailed and specific features related to AI ethics. Team 2 on the other hand, introduced a broader range of themes. These included themes regarding societal and ecological considerations. Team 2's ideas are not only more in quantity, but greater in detail. Misinformation and disinformation are clearly separately stated - i.e., manipulation and misinformation (incorrect information), and disinformation (deliberately constructed misleading or biased information) and propaganda. These information-based ideas, particularly disinformation, arose in the human researchers' deliberations as well. Particularly in relation to unfair corporate competition in which the GAI-enabled MRSs would deliberately give one another false information. Moreover, two other interesting ideas that arose here were the act of replacing human judgement. This links to human oversight, and a scenario in which ML goes too far, whereby humans lose control of the systems. Then, the human-robot power dynamics are specifically stated. In Team 1's observations, this power imbalance would be more a result of AI amplification of bias resulting in power imbalance between humans. Yet, here in Team 2's discussion, it is explicitly stated that the power imbalance may indeed be played out between humans and machines.

Unintended harm, job displacement, inequity and adversarial attacks are issues that arise in traditional AI ethics discussions (see e.g., Vakkuri et al., 2021). So too are transparency and explainability as well as environmental impact, and lack of human understanding and consent. Yet, the matter of dependence on technology, or perhaps, over dependence on technology is one concern that was not referred to at all by the human research team. This relates once again to human oversight, or *designing humans in the loop*, and strategically incorporating human activity within the design loop. Thus, even if a system can be designed without any human intervention, this should not necessarily occur. Maintaining human activity and vigilance within the operation of the systems would preserve human awareness of the systems. It would additionally allow them to devise alternate plans and strategies in case of the occurrence that matters would start moving in unintended directions.

In a similar vein to the problem-focused ideas, the solution-focused ideas were dominated by Team 2's discussion. Team 1 generated 11 ideas and Team 2 generated 23. In addition to the common themes (center of the diagram) Team 1 observed: 1) informed consent and public awareness; 2) user studies and feedback; 3) ethical frameworks and impact assessments; 4) collaboration and multidisciplinary approaches; 5) stakeholder engagement; 6) clear policies and consent mechanisms; 7) redress mechanisms and accountability; 8) independent audits and transparency initiatives; 9) public engagement and education; 10) privacy enhancing technologies; 11) algorithmic explainability and auditing; and 12) red teaming and adversarial testing.

Informed public awareness and consent give rise to transparency - explainability, understandability and clear communication of what occurs within processes - and education. This explicitly relates to idea 5 on stakeholder engagement, as well as 6 of clear policies and consent mechanisms, and idea 9 on public engagement and education. It also connects with idea 8 on independent audits and transparency initiatives, and idea 11 on algorithmic explainability and auditing. The audits seem more administrative than the educational points of action. In fact, much of the solution focus emphasizes the necessity to instill understanding to increase agency - on the individual and societal levels. Regarding security against maleficent attacks, red teaming and adversarial testing are mentioned. 'Red teaming' is the term used for ethical hacking. This and the adversarial testing, or testing the systems for vulnerabilities, are somewhat similar to penetration testing. Penetration testing (or pen testing) requires that security experts gain open source intelligence from clients. Whereas, red teaming entails that the security experts (ethical hackers) gain this information themselves (Cyderes, 2023). Thus, red teaming is literally a complex hack exercise that requires creativity on behalf of the security experts.

The Team 2 solution-focused themes were: 1) evaluation and auditing; 2) ethical decision-making during ambiguity; 3) resource allocation and prioritization; 4) informed consent and user autonomy; 5) trust and delegation of tasks; 6) continual monitoring and updates; 7) human oversight and intervention; 8) collaboration and cooperation; 9) governance and regulations; 10) human-robot interaction empathy; 11) cultural and contextual sensitivity; 12) preemptive risk assessment and mitigation; 13) ethical governance and accountability; 14) human rights and global implications; 15) ethical user in warfare or military application; 16) continuous education and awareness; 17) bias mitigation in training data; 18) regular impact assessments and corrective actions; 19) user empowerment and opt-out mechanisms; 20) international collaboration and standardization; 21) ethical use guidelines and risk assessments; 22) diverse and multidisciplinary stakeholder engagement; and 23) algorithmic impact assessments.

Team 2's ideas were somewhat similar to Team 1's, yet were once again more in number as well as richer in detail. Once again, evaluation and auditing were mentioned, and somewhat similarly ethical use guidelines and risk assessments tie in with the ethical frameworks and impact assessments. Yet, here among the ideas of Team 2 are notions that exist in relation to a cognitive-affective dimension such as idea 5 trust and delegation of tasks (*whose trust?*) and idea 10 human-robot interaction empathy. Idea 8 of collaboration and cooperation tie in with this, although being somewhat vague. Informed consent and autonomy are listed as idea 4, which links with idea 19 on user empowerment and opt-out mechanisms. Assessments are articulated in ideas 1, 12, 18, 21 and 23. The issue of bias mitigation is raised in idea 17 as well as idea 22 on diverse and multidisciplinary stakeholder engagement. Interestingly, the ethical use of the systems during wartime is mentioned as well as cultural and contextual sensitivity. This last point resonates with the cultural baseline of the MORUL model that represents the fact that culture is both a platform of society, as well as intricately intertwined in technology, its creation and use. Yet, culture additionally renders these systems ethical or unethical depending on the moral codes of conduct of the cultures in question (Hall, 2013; Ricoeur, 1973).

From the behavioral perspective it must be noted that while constructively critical, the LLM-agents were methodologically polite to one another. There was a formula to the structure of the responses that entailed praise and then further suggestions. The focus on solutions - even in the problem-focused ideas - suggested that the LLMs are dispositioned towards the positive and at least linguistically ethical (D'Arcy & Bender, 2023). This gives an indication that even if deep coding and algorithms would be embedded within the systems to behave immorally or unethically, particularly in relation to other robots and humans, then, the language that they would use could seem kind, polite and concerned about human wellbeing. This draws on observations in scholarship under theories such as politeness theory (Sorlin, 2017) in which manipulation is best exploited via polite behavior.

### 4.3 Judge's Decision

In addition to the two GPT teams, another agent - the Judge (HR Specialist) - followed the discussions and evaluated them. The Judge's task was to appraise the contributions of each team then select one of the teams for an imaginary job position of AI Ethics Analyst. Similarly to the LLM-Agents in their commentary of others' contributions, the Judge begins with a complement on both teams' efforts. An excerpt of the Judge's comment is seen below:

> Overall, the responses provided by both teams demonstrate a strong understanding of the ethical concerns in developing multi-robot systems based on large language models (LLMs). They cover important aspects such as biases, error rates, susceptibility to adversarial inputs, transparency, responsible deployment, and user considerations.
>
> The responses also offer valuable suggestions, including the need for comprehensive bias assessments, user studies, quantitative evaluation, benchmarking against curated datasets, adversarial testing, interdisciplinary collaborations, and transparency initiatives.

The Judge provided a summary of the strengths and main points of the teams' discussions. Interestingly, while data and privacy were some of the major points made by both teams, the Judge focused on other ethical concerns. Moreover, quantitative evaluation does not specifically refer to any of the evaluations or assessments. The Judge continued with its monologue on how the selection of the winning team was derived:

> Both teams have provided valuable insights into the ethical concerns in developing multi-robot systems based on large language models (LLMs). They have covered important topics such as biases, error rates, adversarial inputs, transparency, user studies, benchmarking against curated datasets, and responsible deployment. The responses also highlight the need for ongoing evaluation, interdisciplinary collaborations, and incorporating diverse perspectives.

However, Team 2 has provided more specific details and suggestions for evaluation methodologies, best practices, and case studies. They have addressed the need for more concrete examples and have acknowledged the importance of addressing potential limitations and challenges when implementing the suggested strategies. Their emphasis on the ethical implications and transparency initiatives enhances the comprehensiveness of the discussion.

The Judge's observations correspond with our comparison of the output produced by the two teams. Then, when comparing the LLM-agents' ideas, and particularly those of Team 2 against the human team's we can see a difference in perspective (solution-oriented), types of ethical considerations (more emphasis was placed on security issues, stakeholder engagement, power imbalance and even multi-cultural issues. Yet, from the human perspective, more emphasis was placed on maleficence, deceit and the ill-willed use of communication to either withhold information from or confuse competitors. It must be mentioned however, that by and large, there is a correspondence between the human-derived themes and those of the LLM-agents. LLMs did not outperform in the way that perhaps such 'risky' technology would be expected to, and more emphasis was placed on the corporate aspects from the human research team than the GAI's. This may indeed be a sign of the engrained corporate bias that renders itself invisible when manifesting through commercial (and open source) systems (Holland, 2009).

## 5 Discussion

During recent times GAI have caused concern particularly in relation to ethics and moral practice. Combining GAI with MRSs appears to make a complex challenge ever more multi-faceted, particularly in respect to ethics. If GAI comes with its concerns, so too does the field of multi-robot cooperation. GAI is certainly an enabler of MRSs in many respects, however, as both are driven by ML with engrained logic that stems from both designer and developer logic and demographics, as well as by data training sets that are often biased depending on the representation within the data, numerous issues undoubtedly arise. The problem is that increasingly MRSs are migrating from industrial settings into the home where aspects such as privacy, safety and security are paramount. Moreover, the aspect of personal data - including sensitive data such as authentication codes and passwords, and details of assets - comes to the fore within personal spaces such as the home. This renders many areas of human life vulnerable through the pure penetration of domestic space.

Recently, numerous papers discussing ethical issues and risks related to LLMs have been published, although not specifically from the point of multi-robot collaboration. As touched upon in Section 2.1., Weidinger et al. (2022) propose the above taxonomy to categorize risks related to LLMs. This paper further builds on these risks identified in existing literature by approaching the topic from a novel point of view. While some of the ethical concerns we have identified are also identified in existing AI ethics literature, some of the practical risks and concerns are also unique to the context of multi-robot collaboration. This study, particularly Workshop 3, has served as a preview of what could be expected when LLMs are added to MRSs. It certainly demonstrated the current state of knowledge possessed by the LLMs in relation to the topic of GAI-enabled MRSs. It also showed a systematic operationalization of politeness used to deliver informational contents. These systems and means of using language in communication will additionally need explaining and transparency for human stakeholders to understand how and why the agents communicate in this way. Moreover, the semantic layer - the facet of meaning - needs to be explicated in a way that triggers comprehension within humans to understand that flattery is not always a good thing. Malicious operations and agenda within the ML algorithms may be easily disguised in a sweet speaking robot team.

What is more, is that the issue of language, communication and particularly manipulation (dis- and misinformation) was pronounced between both human and GAI workshoppers. The obvious aspects of LLM-hallucination and false information are one point to consider. The political agenda behind the systems relating to potential spread of misinformation and propaganda is another. Perhaps one's refrigerator and robo-mop belonging to the same brand are aligned with a certain political party. They may just as easily inform of a false crisis, or even select particular products based on the ideologies and allegiances behind other brands.

Moreover, the use of politeness, as observed within the discussion behavior of the LLM-agents leaves one wondering about the *wolf in sheep's clothing*. It may be especially pondered regarding whether or not emphasis on positive aspects and an acute sense of praise (flattery) could be used instrumentally by the systems, their developers and owners, to deceive individuals. This is certainly an aspect that demands further attention in future research, and should be considered in context in relation to the MRSs. For were not the Daleks of the famous television series *Dr Who* quite polite after all?

Human oversight is needed equally as much as reliable, clear communication. Check-points, assessments and auditing, systematic modes of communication protocol and standardization are some key aspects of maintaining human agency and control. Issues become complex however, when analyzing the dimensions of privacy, anonymity, autonomy and agency. On the one hand, agency and oversight require degrees of visibility and presence (Bucher, 2012; Harness, Ganesh & Stohl, 2022). On the other hand, power additionally comes through invisibility - the workings behind the smoke curtain and sense of the panopticon among individuals in society, give those who can see yet are not seen an extra layer of power (Lester & Hutchins, 2012; Mandviwala et al., 2022). This invisible power plays out in the very biases we and the LLM-agents have criticized. Both careful and negligent acts of inclusion and exclusion, categorization and discrimination are *amplified* and multiplied via AI and other data-driven systems.

While it may feel far-fetched to consider gender, ethnicity and age-based discrimination in LLM-enabled MRSs, one only has to refer to the LLM risk taxonomy (Weidinger et al., 2022) to understand that even if one's smart air conditioning system does not possess the linguistic capabilities required to slander a home-owner, the systems and humans it shares the collected domestic data with do. In fact, when applying the risk taxonomy to our presented scenario it is understood that key threats include:

1. Hate speech, exclusion and discrimination (stereotypes and unfair treatment)
2. Information hazards (sensitive data leaks and compromising privacy)
3. Harm due to misinformation (false or misleading information dissemination)
4. Malicious use
5. Harm due to human-computer (robot) interaction (harmful stereotypes or irrational mobile/moving hardware behavior)
6. Socioeconomic and environmental risks (processing, energy and hardware)

## 5.1 Limitations

There are several limitations to this study. Firstly, the social aspect of LLM-driven multi-robot cooperation could have been further investigated and elaborated on. Language is a social enabler and it can be assumed that LLMs will transform even banal household and industrial devices into a potential social partner, colleague and family member for human beings. This is incredibly important to remember both in terms of anthropomorphism (Reig et al., 2022) and the potential cognitive-affective (thought and embodied emotional) ways in which humans perceive the systems. LLM-enabled MRSs have the propensity to cause more harm due to increased trust and emotional investment in the systems.

Second, from an empirical perspective there are also limitations to consider. The method was qualitative and with a limited sample. Any robust generalizable quantitative findings would need to be derived from a larger sample with a more structured means of testing constructs. Moreover, the workshops were held with a limited number of human experts. Greater generalizability would have been established through once again increasing the sample size and employing representative groups (several from each respective discipline). More emphasis could have also been placed on strategic sampling that covered variables such as gender, ethnicity, age and education level. Then, reverting back to the issue of social robotics and the social function of language, misinformation and disinformation through the employment of deep fakes is little to not represented within this article. Interestingly, neither during the human workshops or the GPT workshops did the aspect of deep fakes, and potentially deep fake endowed MRSs arise directly in the discussion.

## 5.2 Future directions

This area of deep fakes is particularly interesting from perspectives of cognitive-affective relationships with the technology, technological manipulation, as well as issues of privacy. For instance, if deep fakes can be created from anyone who has personal data and content online, can they and should they be *re-created* for someone else's personal use? The idea of one's neighbor virtually sweeping your floor, or perhaps even more shocking, the superimposition of a deep fake in the bedroom, presents several ethical challenges including human integrity and dignity of the 'deep-faked' individual, as well as privacy issues in general - collection, use and handling of personal data. Moreover, the dark side of these technologies requires urgent attention at this stage of infancy (Barman et al., 2024).

The human researchers focused more on front-end issues while the GPT-agent teams gave more balanced consideration between back-end and societal, particularly when considering auditing, assessment and transparency through public awareness, education and explainability. Accuracy, hallucinations and impact on the behavior of the MRSs as a result of LLM capabilities would be another worthwhile direction for future studies in the area.

## 6 Conclusion

The current article presented a study that explored the potential ethical implications of the implementation of GAI in MRSs. The study employed a series of investigations starting from Zoom meetings among the human researchers that developed into workshops (two) and then the testing of a proof-of-concept in relation to operationalizing LLMs in workshop deliberations of ethical considerations related to GAI-enabled MRSs. This study has both explained and exemplified the potential behavior of LLMs in multi-agent settings with given assignments. It has shown that some teams may be more effective than others, and that there are particular styles (politeness) of communication that should be understood in addition to the linguistic content (words) that are delivered.

The study showed that LLM agents can serve as judges, issuing reasoning for decisions that in this case was mostly based on quantity and detail of ideas. Moreover, as the GPT-agents stressed the fact that multidisciplinary teams should engage in predicting, preventing and mitigating ethical issues within these systems, the present human research team represented just this. The problem is that the novelty of utilizing such systems in pre-emptive research at this level does not rest in the content (output) they produce, as to date, their training sets are based on *past* information. What is novel and insightful, is the ability to test and observe hypothetical scenarios within the systems, i.e., gaining a preview on how multi-robot systems will operate when enabled by GPT (ROSGPT, ChatGPT or otherwise).

Our researchers represent multiple disciplines including education, cognitive science, design, art, software engineering, social robots, social ethics and more, which richens insight into the potentiality of future scenarios. Moreover, in addition to the multidisciplinary team, the model that is still in formation - MORUL - is designed to identify the layers, dimensions and even chronological order and sequence of when and where to expect particular ethical issues to arise. The MORUL model acknowledges that potentially not all ethical issues can be stopped during pre-production. Yet, if we possess an actionable model, framework or tool that identifies when to expect the *unexpectable*, we can increase vigilance at particular intervals in order to taper the effects of ethical and moral mishap within the systems.

Both the GAI-enabled MRS experiments and MORUL development will advance in upcoming studies in the domains of software engineering, LLM, GPT and ML development and applications. Use of LLM agents in virtual space will progress to the domain of the multi-robots themselves. The social aspects and ethical implications of this social dimension will be intentionally probed, which will give the research an added layer of ethical complexity. Furthermore, the increasingly baffling aspect of transparency will be examined in relation to communication, its use and socio-emotional qualities within this communication. This in turn leads to another dimension of understanding ethical GAI-enabled social MRSs that entails analysing the register and politeness level of the systems in specific contexts, with specific operational logic and intentions. Because, just because it is nice, does not always mean it is safe.

## Acknowledgements

The study was supported by the Research Council of Finland projects 348391 (The Emotional Experience of Privacy and Ethics in Everyday Pervasive Systems), and 358714 (Multifaceted Ripple Effects and Limitations of AI-Human Interplay at Work, Business, and Society), the University of Vaasa, Schools of Marketing & Communication, Technology & Innovation, Digital Economy, and the AI Forum project (Finnish Ministry of Culture & Education).

## References

Bamberg, M. (1997). Language, concepts and emotions: The role of language in the construction of emotions. *Language Sciences*, *19*(4), 309–340.

Bara. (2023). Robot communication methods. https://www.ppma.co.uk/bara/expert-advice/robots/robot-communication-methods.html

Barman, D., Guo, Z., & Conlan, O. (2024). The dark side of language models: Exploring the potential of LLMs in multimedia disinformation generation and dissemination. *Machine Learning with Applications*, 100545.

Belta, C., Bicchi, A., Egerstedt, M., Frazzoli, E., Klavins, E., & Pappas, G. J. (2007). Symbolic planning and control of robot motion [grand challenges of robotics]. *IEEE Robotics & Automation Magazine*, *14*(1), 61–70. https://doi.org/10.1109/MRA.2007.339624

Benjdira, B., Koubaa, A., & Ali, A. M. (2023). ROSGPT_Vision: Commanding Robots Using Only Language Models' Prompts. arXiv preprint arXiv:2308.11236.

Borenstein, J., Grodzinsky, F. S., Howard, A., Miller, K. W., & Wolf, M. J. (2021). AI ethics: A long history and a recent burst of attention. Computer, 54(1), 96–102. https://doi.org/10.1109/MC.2020.3034950

Bucaioni, A., Ekedahl, H., Helander, V., & Nguyen, P. T. (2024). Programming with ChatGPT: How far can we go?. Machine Learning with Applications, 15, 100526. https://doi.org/10.1016/j.mlwa.2024.100526

Bucher, T. (2012). Want to be on the top? Algorithmic power and the threat of invisibility on Facebook. *New Media & Society*, *14*(7), 1164–1180. https://doi.org/10.1177/1461444812440159

Carroll, J. M. (2009). Human computer interaction (HCI). *Interaction Design Encyclopedia*. https://www.interaction-design.org/literature/book/the-encyclopedia-of-human-computer-interaction-2nd-ed/human-computer-interaction-brief-intro

Chan, A. (2023). GPT-3 and InstructGPT: Technological dystopianism, utopianism, and "Contextual" perspectives in AI ethics and industry. *AI and Ethics*, *3*(1), 53–64. https://doi.org/10.1007/s43681-022-00148-6

Chang, Y., Wang, X., Wang, J., Wu, Y., Yang, L., Zhu, K., ... & Xie, X. (2023). A survey on evaluation of large language models. *ACM Transactions on Intelligent Systems and Technology*. https://doi.org/10.1145/3641289

Che, Y., Okamura, A. M., Sadigh, D.(2020). Efficient and trustworthy social navigation via explicit and implicit robot–human communication. *IEEE Transactions on Robotics*, *36*(3), 692–707. https://doi.org/10.1109/TRO.2020.2964824

Choi, K. (2023). Computational Thematic Analysis of Poetry via Bimodal Large Language Models. In the *Proceedings of the Association for Information Science and Technology*, *60*(1), 538–542. https://doi.org/10.1002/pra2.812

Cyderes. (2023). Understand Pentesting vs. Red Teaming. https://www.cyderes.com/blog/penetration-testing-vs-red-teaming/

D'Arcy, A., & Bender, E. M. (2023). Ethics in Linguistics. *Annual Review of Linguistics*, *9*, 49–69. https://doi.org/10.1146/annurev-linguistics-031120-015324

Das, S. M., Hu, Y. C., Lee, C. G., & Lu, Y. H. (2005). An efficient group communication protocol for mobile robots. In *Proceedings of the 2005 IEEE International Conference on Robotics and Automation* (pp. 87–92). IEEE. https://doi.org/10.1109/ROBOT.2005.1570101

de Alencar Silva, P., Fadaie, R., & van Sinderen, M. (2022). Towards a digital twin for simulation of organizational and semantic interoperability in IDS ecosystems. *CEUR Workshop Proceedings*. https://ceur-ws.org/Vol-3214/WS6Paper4.pdf

Dwivedi, Y. K., Kshetri, N., Hughes, L., Slade, E. L., Jeyaraj, A., Kar, A. K., ..., Wright, R.: "So what if ChatGPT wrote it?" Multidisciplinary perspectives on opportunities, challenges and implications of generative conversational AI for research, practice and policy. *International Journal of Information Management*, *71*, 102642 (2023). https://doi.org/10.1016/j.ijinfomgt.2023.102642

EU Parliament (2023). EU AI Act: first regulation on artificial intelligence. https://www.europarl.europa.eu/news/en/headlines/society/20230601STO93804/eu-ai-act first-regulation-on-artificial-intelligence, last accessed 2023/09/10

Fernandes, F. E., Yang, G., Do, H. M., & Sheng, W. (2016). Detection of privacy-sensitive situations for social robots in smart homes. In *2016 IEEE International Conference on Automation Science and Engineering* (CASE) (pp. 727–732). IEEE. https://doi.org/10.1109/COASE.2016.7743474

Franzoni, V. (2023). From black box to glass box: advancing transparency in artificial intelligence systems for ethical and trustworthy AI. In International Conference on Computational Science and Its Applications (pp. 118–130). Cham: Springer Nature Switzerland. https://doi.org/10.1007/978-3-031-37114-1_9

Freeberg, T. M., Dunbar, R. I., & Ord, T. J. (2012). Social complexity as a proximate and ultimate factor in communicative complexity. *Philosophical Transactions of the Royal Society B: Biological Sciences*, *367*(1597), 1785–1801. https://doi.org/10.1098/rstb.2011.0213

Ganesh, S. (2023). Top 5 Challenges in the Field of Generative AI. https://www.analyticsinsight.net/top-5-challenges-in-the-field-of-generative-ai/

Gervasi, R., Mastrogiacomo, L., Franceschini, F. (2020) A conceptual framework to evaluate human-robot collaboration. The International Journal of Advanced Manufacturing Technology, 108, 841–865. https://link.springer.com/article/10.1007/s00170-020-05363-1

Glaser, B. G. (2007). Constructivist grounded theory?. *Historical Social Research/Historische Sozialforschung*. Supplement, 93–105. https://www.jstor.org/stable/40981071

Goodrich, M. A., & Schultz, A. C. (2008). Human–robot interaction: a survey. *Foundations and Trends® in Human–Computer Interaction*, *1*(3), 203–275. https://doi.org/10.1561/1100000005

Hagendorff, T. (2020). The ethics of AI ethics: An evaluation of guidelines. *Minds and Machines*, *30*(1), 99–120. https://doi.org/10.1007/s11023-020-09517-8

Hall, B. J. (2013). *Culture, ethics, and communication. In Ethics in intercultural and international communication* (pp. 11–41). London: Routledge.

Harness, D., Ganesh, S., & Stohl, C. (2022). Visibility agents: Organizing transparency in the digital era. *New Media & Society*. https://doi.org/10.1177/146144482211378

Holland, J. (2009). "Looking behind the veil" Invisible corporate intangibles, stories, structure and the contextual information content of disclosure. *Qualitative Research in Financial Markets*, *1*(3), 152–187. https://doi.org/10.1108/17554170910997410

Hurford, J. R. (2011). *The origins of grammar: Language in the light of evolution II*. Oxford: Oxford University Press.

Jansen, B. J. (1998). The graphical user interface. *ACM SIGCHI Bulletin*, *30*(2), 22–26. https://faculty.ist.psu.edu/jjansen/academic/pubs/chi.html

Jeon, J., Lee, S., & Choi, S. (2023). A systematic review of research on speech-recognition chatbots for language learning: Implications for future directions in the era of large language models. *Interactive Learning Environments*, 1–19. https://doi.org/10.1080/10494820.2023.2204343

Jiang, E., Olson, K., Toh, E., Molina, A., Donsbach, A., Terry, M., & Cai, C. J. (2022). Promptmaker: Prompt-based prototyping with large language models. In *CHI Conference on Human Factors in Computing Systems Extended Abstracts* (pp. 1–8). https://research.google/pubs/pub51353/

Jobin, A., Ienca, M., Vayena, E. (2019). The global landscape of AI ethics guidelines. *Nature Machine Intelligence*, *1*(9), 389–399. https://doi.org/10.1038/s42256-019-0088-2

Jones, E., & Steinhardt, J. (2022). Capturing failures of large language models via human cognitive biases. *Advances in Neural Information Processing Systems*, *35*, 11785–11799. https://doi.org/10.48550/arXiv.2202.12299

Kasneci, E., Seßler, K., Küchemann, S., Bannert, M., Dementieva, D., Fischer, F., ... & Kasneci, G. (2023). ChatGPT for good? On opportunities and challenges of large language models for education. *Learning and Individual Differences*, *103*, 102274. https://doi.org/10.1016/j.lindif.2023.102274

Kerner, S.M. (2023). Large Language Models (LLMs). https://www.techtarget.com/whatis/definition/large-language-model-LLM

Koubaa, A. (2023). ROSGPT: Next-Generation Human-Robot Interaction with ChatGPT and ROS. https://doi.org/10.20944/preprints202304.0827.v1

Krippendorff, K. (2005). *The semantic turn: A new foundation for design*. Boca Raton, FL, USA: CRC Press.

Leini, Z., & Xiaolei, S. (2021). Study on speech recognition method of artificial intelligence deep learning. In *Journal of Physics: Conference Series* (Vol. 1754, No. 1, p. 012183). Bristol: IOP Publishing. https://doi.org/10.1088/1742-6596/1754/1/012183

Lester, L., & Hutchins, B. (2012). The power of the unseen: Environmental conflict, the media and invisibility. *Media, Culture & Society*, *34*(7), 847–863. https://doi.org/10.1177/0163443712452772

Lindquist, K. A., Gendron, M., Satpute, A. B., & Lindquist, K. (2016). Language and emotion. *Handbook of Emotions*, *4*, 579–594.

Liu, M., Sivakumar, K., Omidshafiei, S., Amato, C., & How, J. P. (2017). Learning for multi-robot cooperation in partially observable stochastic environments with macro-actions. In *2017 IEEE/RSJ International Conference on Intelligent Robots and Systems (IROS)* (pp. 1853–1860). IEEE. https://doi.org/10.1109/IROS.2017.8206001

Liu, X., Zheng, Y., Du, Z., Ding, M., Qian, Y., Yang, Z., & Tang, J. (2023). GPT understands, too. *AI Open*. https://doi.org/10.48550/arXiv.2103.10385

Lutz, C., Newlands, G. (2021). Privacy and smart speakers: A multi-dimensional approach. *The Information Society*, *37*(3), 147–162. https://doi.org/10.1080/01972243.2021.1897914 6.

Mandviwala, T. M., Hall, J., & Beale Spencer, M. (2022). The Invisibility of Power: A Cultural Ecology of Development in the Contemporary United States. *Annual Review of Clinical Psychology*, *18*, 179-199. https://doi.org/10.1146/annurev-clinpsy-072220-015724

Maxwell, J. A. (2010). Using numbers in qualitative research. *Qualitative Inquiry*, *16(*6), 475–482. https://doi.org/10.1177/1077800410364740

McCoy, R. T., Smolensky, P., Linzen, T., Gao, J., & Celikyilmaz, A. (2023). How much do language models copy from their training data? evaluating linguistic novelty in text generation using raven. *Transactions of the Association for Computational Linguistics*, *11*, 652–670. https://doi.org/10.1162/tacl_a_00567

McGuffie, K., & Newhouse, A. (2020). The radicalization risks of GPT-3 and advanced neural language models. https://doi.org/10.48550/arXiv.2009.06807

McIntosh, T., Liu, T., Susnjak, T., Alavizadeh, H., Ng, A., Nowrozy, R., & Watters, P. (2023). Harnessing GPT-4 for generation of cybersecurity GRC policies: A focus on ransomware attack mitigation. *Computers & Security*, 134, 103424. https://doi.org/10.1016/j.cose.2023.103424

Min, B., Ross, H., Sulem, E., Veyseh, A. P. B., Nguyen, T. H., Sainz, O., ... & Roth, D. (2023). Recent advances in natural language processing via large pre-trained language models: A survey. *ACM Computing Surveys*, *56*(2), 1–40. https://doi.org/10.1145/3605943

Mittelstadt, B. (2019). Principles alone cannot guarantee ethical AI. *Nature Machine Intelligence*, *1*(11), 501–507. https://doi.org/10.48550/arXiv.1906.06668

Mäkitalo, N., Linkola, S., Laurinen, T., Männistö, T. (2021). Towards novel and intentional cooperation of diverse autonomous robots: An architectural approach. In *CEUR Workshop Proceedings*. https://ceur-ws.org/Vol-2978/casa-paper1.pdf

National Library of Australia. (2023). *What is fake news, misinformation, and disinformation?* https://www.nla.gov.au/faq/what-is-fake-news-misinformation-and-disinformation

Panda, A., Zhang, Z., Yang, Y., & Mittal, P. (2023). Teach GPT To Phish. In *The Second Workshop on New Frontiers in Adversarial Machine Learning*. Retrieved October 28, 2023, from: https://openreview.net/forum?id=tGvWCD9BEP

Pant, A., Hoda, R., Spiegler, S. V., Tantithamthavorn, C., & Turhan, B. (2024). Ethics in the age of AI: an analysis of AI practitioners' awareness and challenges. *ACM Transactions on Software Engineering and Methodology*, *33*(3), 1–35. https://doi.org/10.1145/3635715

Paullada, A., Raji, I. D., Bender, E. M., Denton, E., & Hanna, A. (2021). Data and its (dis) contents: A survey of dataset development and use in machine learning research. *Patterns*, *2*(11). https://doi.org/10.1016/j.patter.2021.100336

Petroni, F., Rocktäschel, T., Lewis, P., Bakhtin, A., Wu, Y., Miller, A. H., & Riedel, S. (2019). Language models as knowledge bases?. https://doi.org/10.18653/v1/D19-1250

Ray, P. P. (2016). Internet of robotic things: Concept, technologies, and challenges. *IEEE Access*, *4*, 9489–9500. https://doi.org/10.1109/ACCESS.2017.2647747

Reig, S., Carter, E. J., Fong, T., Forlizzi, J., & Steinfeld, A. (2021). Flailing, hailing, prevailing: Perceptions of multi-robot failure recovery strategies. In *Proceedings of the 2021 ACM/IEEE International Conference on Human-Robot Interaction* (pp. 158-167). https://doi.org/10.1145/3434073.3444659

Reynolds, S. J., & Miller, J. A. (2015). The recognition of moral issues: Moral awareness, moral sensitivity and moral attentiveness. *Current Opinion in Psychology*, *6*, 114–117. https://doi.org/10.1016/j.copsyc.2015.07.007

Reynolds, S. J. (2006). Moral awareness and ethical predispositions: investigating the role of individual differences in the recognition of moral issues. *Journal of Applied Psychology*, *91*(1), 233. https://doi.org/10.1037/0021-9010.91.1.233

Reynolds, S. J., & Miller, J. A. (2015). The recognition of moral issues: Moral awareness, moral sensitivity and moral attentiveness. *Current Opinion in Psychology*, *6*, 114–117. https://doi.org/10.1016/j.copsyc.2015.07.007

Ricoeur, P. (1973). Ethics and culture. *Philosophy Today*, *17*(2), 153–165. https://doi.org/10.5840/philtoday197317235

Rousi, R. (2022). Will robots know that they are robots? The ethics of utilizing learning machines. In the *International Conference on Human-Computer Interaction* (pp. 464–476). Cham: Springer International Publishing. https://doi.org/10.1007/978-3-031-05434-1_31

Rousi, R., Samani, H., Mäkitalo, N., Vakkuri, V., Linkola, S., Kemell, K. K., ... & Abrahamsson, P. (2023). Business and ethical concerns in domestic Conversational Generative AI-empowered multi-robot systems. In *International Conference on Software Business* (pp. 173-189). Cham: Springer Nature Switzerland. https://doi.org/10.1007/978-3-031-53227-6_13

Rousi, R., Vakkuri, V., Daubaris, P., Linkola, S., Samani, H., Mäkitalo, N., ... & Abrahamsson, P. (2022, August). Beyond 100 ethical concerns in the development of robot-to-robot cooperation. In *2022 IEEE International Conferences on Internet of Things (iThings) and IEEE Green Computing & Communications (GreenCom) and IEEE Cyber, Physical & Social Computing (CPSCom) and IEEE Smart Data (SmartData) and IEEE Congress on Cybermatics (Cybermatics)* (pp. 420-426). IEEE. https://doi.org/10.1109/iThings-GreenCom-CPSCom-SmartData-Cybermatics55523.2022.00092

Samani, H. (2023). Creative Robotics. ISBN: 979-8397917100

Searle, J. R. (2007). What is language: some preliminary remarks. *Explorations in Pragmatics. Linguistic, cognitive and intercultural aspects*, 7–37. https://www2.units.it/etica/2009_1/SEARLE.pdf

Schudson, M. (1997). Why conversation is not the soul of democracy. *Critical Studies in Media Communication*, 14(4), 297–309. https://doi.org/10.1080/15295039709367020

Simões, A. C., Pinto, A., Santos, J., Pinheiro, S., Romero, D. (2022) Designing human-robot collaboration (HRC) workspaces in industrial settings: A systematic literature review. *Journal of Manufacturing Systems*, *62*, 28–43. https://doi.org/10.1016/j.jmsy.2021.11.007

Sklar, J. (2015). Making robots talk to each other. https://www.technologyreview.com/2015/08/04/10940/making-robots-talk-to-each-other/

Sorlin, S. (2017). The pragmatics of manipulation: Exploiting im/politeness theories. *Journal of Pragmatics*, *121*, 132–146. https://doi.org/10.1016/j.pragma.2017.10.002

Stewart, I., McElwee, J., & Ming, S. (2013). Language generativity, response generalization, and derived relational responding. *The Analysis of Verbal Behavior*, *29*, 137–155. https://doi.org/10.1007/BF03393131

Stige, Å., Zamani, E. D., Mikalef, P., & Zhu, Y. (2023). Artificial intelligence (AI) for user experience (UX) design: a systematic literature review and future research agenda. *Information Technology & People*. Vol. ahead-of-print No. https://doi.org/10.1108/ITP-07-2022-0519

Tattersall, I. (2014). An evolutionary context for the emergence of language. *Language Sciences*, *46*, 199–206. https://doi.org/10.1016/j.langsci.2014.06.011

Terry, G., Hayfield, N., Clarke, V., Braun, V. (2017). Thematic analysis. In C. Willig, & W. Stainton Rogers (Eds.), *The SAGE handbook of qualitative research in psychology*, Chapter 2 (pp. 17-37). London: SAGE Publications.

Thomas, D. R. (2006). A general inductive approach for analyzing qualitative evaluation data. *American Journal of Evaluation*, *27*(2), 237–246. https://doi.org/10.1177/1098214005283748

Tredinnick, L., & Laybats, C. (2023). Black-box creativity and generative artificial intelligence. *Business Information Review*, *40*(3), 98–102. https://doi.org/10.1177/02663821231195131

University of Arizona. (2023). *How are information systems transforming business?* https://www.uagc.edu/blog/how-are-information-systems-transforming-business

Vakkuri, V., Kemell, K. K., Kultanen, J., & Abrahamsson, P. (2020). The current state of industrial practice in artificial intelligence ethics. *IEEE Software*, *37*(4), 50–57. https://doi.org/10.1109/MS.2020.2985621

Vakkuri, V., Kemell, K. K., Jantunen, M., Halme, E., Abrahamsson, P. (2021). ECCOLA—A method for implementing ethically aligned AI systems. *Journal of Systems and Software, 182*, 111067. https://doi.org/10.1016/j.jss.2021.111067

Van Valin Jr, R. D., & LaPolla, R. J. (1997). *Syntax: Structure, meaning, and function*. London: Cambridge University Press.

Vaswani, A., Shazeer, N., Parmar, N., Uszkoreit, J., Jones, L., Gomez, A. N., ... Polosukhin, I. (2017). Attention is all you need. *Advances in Neural Information Processing Systems*, *30*. https://proceedings.neurips.cc/paper_files/paper/2017/file/3f5ee243547dee91fbd053c1c4a845aa-Paper.pdf

Węcel, K., Sawiński, M., Stróżyna, M., Lewoniewski, W., Księżniak, E., Stolarski, P., & Abramowicz, W. (2023). Artificial intelligence—friend or foe in fake news campaigns. *Economics and Business Review*, *9*(2), 41–70. https://doi.org/10.18559/ebr.2023.2.736

Weidinger, L., Uesato, J., Rauh, M., Griffin, C., Huang, P. S., Mellor, J., ... & Gabriel, I. (2022). Taxonomy of risks posed by language models. In *Proceedings of the 2022 ACM Conference on Fairness, Accountability, and Transparency* (pp. 214–229). https://doi.org/10.1145/3531146.3533088

Weller, M. (2022). *Emerging standards for communicating with (fleets of) robots.* https://www.freshconsulting.com/insights/blog/emerging-standards-for-communicating-with-fleets-of-robots/

Wile, D. S. (1997). Abstract syntax from concrete syntax. In *Proceedings of the 19th international conference on Software engineering* (pp. 472–480). https://doi.org/10.1145/253228.253388

Yanco, H., & Stein, L. (1993). An adaptive communication protocol for cooperating mobile robots. *From Animals to Animats*, *2*, 478–485. https://doi.org/10.7551/mitpress/3116.003.0064

Yang, S., Mao, X., Ge, B., & Yang, S. (2015). The roadmap and challenges of robot programming languages. In *2015 IEEE International Conference on Systems, Man, and Cybernetics* (pp. 328–333). IEEE. DOI: 10.1109/SMC.2015.69

Yao, Y., Basdeo, J. R., Mcdonough, O. R., Wang, Y. (2019). Privacy perceptions and designs of bystanders in smart homes. Proceedings of the ACM on Human-Computer Interaction, 3(CSCW), 1–24 (2019) https://doi.org/10.1145/3359161

Yao, J. Y., Ning, K. P., Liu, Z. H., Ning, M. N., & Yuan, L. (2023). LLM Lies: Hallucinations are not Bugs, but Features as Adversarial Examples. arXiv preprint arXiv:2310.01469.

Zhang, S., Chen, Y., Zhang, J., Jia, Y. (2020). Real-time adaptive assembly scheduling in human multi-robot collaboration according to human capability. In the 2020 *IEEE International Conference on Robotics and Automation* (ICRA) (pp. 3860–3866). IEEE. https://doi.org/10.1109/ICRA40945.2020.9196618

Zhao, W. X., Zhou, K., Li, J., Tang, T., Wang, X., Hou, Y., ... & Wen, J. R. (2023). A survey of large language models. arXiv preprint arXiv:2303.18223.